\documentclass[10pt,twocolumn,letterpaper]{article}

\usepackage[pagenumbers]{cvpr} 

\usepackage{siunitx}
\usepackage{epsfig}
\usepackage{graphicx}
\usepackage{amsmath,amsthm,amssymb,amstext}
\usepackage{wasysym}
\usepackage{mathtools}
\usepackage{booktabs}
\usepackage{multirow}
\usepackage{balance}
\usepackage{lipsum}
\usepackage{caption}
\usepackage{algorithm}
\usepackage{algorithmic}
\usepackage{enumitem}
\usepackage{engord}
\usepackage{bm}
\usepackage{bbm}
\usepackage{comment}
\usepackage{color}
\usepackage[x11names,dvipsnames,table]{xcolor}
\usepackage{ctable}
\usepackage{setspace}
\usepackage{pifont}
\usepackage{dsfont} 
\usepackage[toc,page]{appendix}
\usepackage{hhline}
\usepackage{listings}
\usepackage{longtable}

\definecolor{cvprblue}{rgb}{0.21,0.49,0.74}
\usepackage[pagebackref,breaklinks,colorlinks,citecolor=cvprblue]{hyperref}

\usepackage[capitalize]{cleveref}

\newcommand\supp{Sup. Mat\xspace}

\definecolor{Gray}{gray}{0.9}
\definecolor{White}{gray}{1}
\definecolor{DGray}{gray}{0.8}
\definecolor{DDDDGray}{gray}{0.3}
\definecolor{WhiteGray}{rgb}{0.9, 0.9, 0.9}
\definecolor{citecolor}{HTML}{0071bc}
\definecolor{darkred}{rgb}{0.6, 0.1, 0.05}
\definecolor{DeltaColor}{rgb}{0.039,0.73,0.71}
\definecolor{SigmaColor}{rgb}{0.98,0.45,0.0}
\definecolor{AlphaColor}{rgb}{0,0,0.8}
\definecolor{BetaColor}{rgb}{0.8,0,0.8}
\definecolor{GammaColor}{rgb}{0.514,0.34,0.224}
\definecolor{EpsilonColor}{rgb}{0.353,0.725,0.906}
\definecolor{PurpleColor}{HTML}{8B008B}
\definecolor{BadColor}{HTML}{C0392B}
\definecolor{OrangeColor}{rgb}{0.914,0.541,0.0.141}
\definecolor{GreenColor}{rgb}{0.137,0.573,0.565}
\definecolor{RedColor}{rgb}{0.949,0.275, 0.224}
\definecolor{LightCyan}{rgb}{0.88,1,1}
\definecolor{DarkCyan}{rgb}{0, 0.502, 0.502}
\definecolor{Gray}{gray}{0.85}
\definecolor{execute}{RGB}{237,28,36}
\definecolor{objtraj}{RGB}{247,147,30}
\definecolor{generator}{RGB}{0,113,188}

\definecolor{bestcolor}{rgb}{1, 0.5, 0.25}
\definecolor{secondbestcolor}{rgb}{1, 0.8, 0.5}

\newcommand{\inlinecode}[1]{\texttt{\small #1}}

\newcommand{\del}[1]{}

\newcommand{\minitab}[2][l]{\begin{tabular}{#1}#2\end{tabular}}
\DeclareMathAlphabet\mathbfcal{OMS}{cmsy}{b}{n}

\newcommand{\cheading}[1]{\noindent\mbox{\textbf{#1}\;}}
\newcommand{\qheading}[1]{\vspace{5pt}\noindent\mbox{\textbf{#1}\;}}

\newcommand{\xmark}{\ding{55}}%
\newcommand{\greencheck}{{\color{Green4} \checkmark}\xspace}
\newcommand{\yellowcheck}{{\color{Goldenrod} \checkmark}\xspace}

\newcommand{\redcross}{{\color{red} \xmark}\xspace}
\newcommand*{\customcircled}[1]{\raisebox{.5pt}{\textcircled{\raisebox{-.9pt} {#1}}}}

\newcommand{\customfootnotetext}[2]{{
  \renewcommand{\thefootnote}{#1}
  \footnotetext[0]{#2}}}

\newcommand{\gt}{\textbf{g.t.}\xspace}
\newcommand{\primitiveid}[1]{\textit{\texttt{#1}}}

\newcommand{\oakinki}{OakInk}
\newcommand{\oakinkiibf}{{\textbf{\textsc{OakInk2}}}}
\newcommand{\oakinkii}{{\textsc{OakInk2}}}

\newcommand{\totalseqnum}{627}
\newcommand{\totalsbjnum}{9}
\newcommand{\totalobjnum}{75}

\newcommand{\totalframenum}{4.01M}

\newcommand{\seqpnum}{363}
\newcommand{\seqcnum}{264}

\newcommand{\targetnum}{38}
\newcommand{\affordancenum}{39}
\newcommand{\ptypenum}{60}
\newcommand{\ctypenum}{150}

\def\paperID{79} 
\def\confName{CVPR}
\def\confYear{2024}

\newcommand{\PaperTitle}{{\textsc{OakInk2} : A Dataset of Bimanual Hands-Object Manipulation \\in Complex Task Completion}}

\title{\PaperTitle}

\author{
    {
        Xinyu Zhan\textsuperscript{1$\star$}\quad
        Lixin Yang\textsuperscript{1$\star$}\quad
        Yifei Zhao\textsuperscript{1}\quad
        Kangrui Mao\textsuperscript{1}\quad
        Hanlin Xu\textsuperscript{1}\quad
    } \\
    {
        Zenan Lin\textsuperscript{2,$\ddagger$}\quad
        Kailin Li\textsuperscript{1}\quad
        Cewu Lu\textsuperscript{1$\boldsymbol{\dagger}$}
    } \\
    {
        \small
        {$^{1}$Shanghai Jiao Tong University, $^{2}$South China University of Technology}
    } 
}

\begin{document}
\newcommand{\teaserCaption}{
\textbf{An overview of the data and content of our proposed \oakinkiibf{} dataset.}
\oakinkii{} dataset focuses on bimanual object manipulation tasks for complex daily activities. \textbf{1) The top row shows the data collection process}, including the task setup (top-left panel), human demonstration (top-center), and annotation (top-right).
\textbf{2) The second row shows the three levels of abstraction constructed by \oakinkii{} for complex tasks}, including the Affordance, Primitive Task, and Complex Task.
\oakinkii{} dataset provides allocentric and egocentric videos of human manipulation process, as well as the corresponding 3D-pose annotation and task specification.
}
\twocolumn[{
    \vspace{-0.0em}
    \maketitle
    \centering
    \vspace{-1.5em}
    \begin{minipage}{1.00\textwidth}
         \centering
         \includegraphics[trim=000mm 000mm 000mm 000mm, clip=False, width=\linewidth]{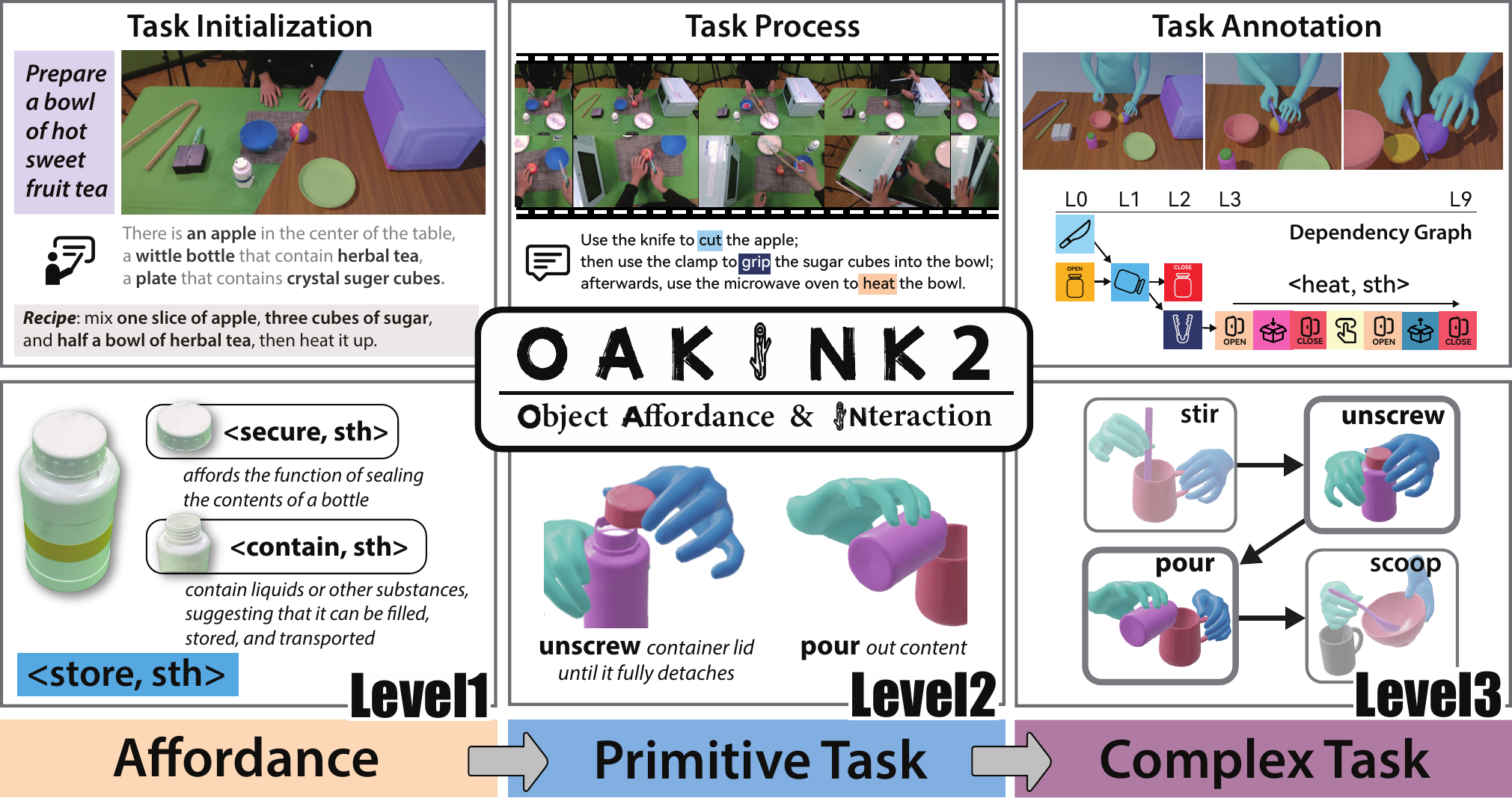}
    \end{minipage}
    \captionsetup{type=figure}
    \captionof{figure}{\teaserCaption}
    \label{fig:teaser}
    \vspace{2.0em}
}]
\customfootnotetext{$\star$}{
    The first two authors contributed equally.
}

\customfootnotetext{$\ddagger$}{
    This work is done when Lin is an intern at SJTU.
}

\customfootnotetext{$\dagger$}{
    Cewu Lu is the corresponding author. He is the member of Qing Yuan Research Institute and MoE Key Lab of Artificial Intelligence, AI Institute, Shanghai Jiao Tong University, China.
}

\begin{abstract}
    We present \oakinkiibf{}, a dataset of bimanual object manipulation tasks for complex daily activities. In pursuit of  constructing the complex tasks into a structured representation, \oakinkii{} introduces three level of abstraction to organize the manipulation tasks: \textbf{Affordance}, \textbf{Primitive Task}, and \textbf{Complex Task}. 
    \oakinkii{} features on an object-centric perspective for decoding the complex tasks, treating 
    them as a sequence of object affordance fulfillment. 
    The first level, Affordance, outlines the functionalities that objects in the scene can afford, the second level, Primitive Task, describes the minimal interaction units that humans interact with the object to achieve its affordance, and the third level, Complex Task, illustrates how Primitive Tasks are composed and interdependent.
    \oakinkii{} dataset provides multi-view image streams and precise pose annotations for the human body, hands and various interacting objects. This extensive collection supports applications such as interaction reconstruction and motion synthesis.
    Based on the 3-level abstraction of \oakinkii{}, we explore a task-oriented framework for Complex Task Completion (CTC). CTC aims to generate a sequence of bimanual manipulation to achieve task objectives. Within the CTC framework, we employ Large Language Models (LLMs) to decompose the complex task objectives into sequences of Primitive Tasks and have developed a Motion Fulfillment Model that generates bimanual hand motion for each Primitive Task.
    \oakinkii{} datasets and models are available at \url{https://oakink.net/v2}.
\end{abstract}
\vspace{-1.0em}

\section{Introduction}

Learning how humans achieve specific task objectives through diverse object manipulation behaviors has been a long-standing challenge. 
Recent data-driven approaches have made significant progress on this topic, including hand-object pose estimation \cite{hasson2019obman,hasson2020leveraging, hasson21homan, rhoi2020, liu2021semi, Doosti2020hopenet, li2021artiboost, hampali2022keypoint, Aboukhadra2023WACV}, interaction synthesis \cite{corona2020ganhand, jiang2021hand, taheri2022goal, wu2022saga, ghosh2023imos, zheng2023cams}, and action imitation \cite{qin2021dexmv, qin2022one}. 
However, the gap still exists for current methods to achieve a human-level understanding on object manipulation for complex task completion.
In particular, humans possess a remarkable capacity to interact with specific objects in an appropriate sequence to achieve desired outcomes \cite{kurby2008segmentation}.
This inspires us to focus on the decomposition of hands-object interaction in complex manipulation tasks into sequential units.

Tracing prior research, the advancement in hand-object interaction understanding is inseparable from the emergence of a series of hand-object interaction datasets \cite{hasson2019obman,corona2020ganhand,hampali2020ho3dv2,brahmbhatt2020contactpose,taheri2020grab,chao2021dexycb,kwon2021h2o,qin2021dexmv,liu2022hoi4d,fan2023arctic,zhu2023contactart,ohkawa2023assemblyhands,jian2023affordpose,yang2022oakink} to support data-driven methods.
A noteworthy example among these datasets is \oakinki{} \cite{yang2022oakink}. 
\oakinki{} analyzed object affordances (\ie functional properties of objects/object-parts \cite{gibson2014ecological}) and collected \textit{human-centric} grasping interaction driven by intents to utilize these affordances. The term: \textit{Oak} is for object affordance knowledge, and \textit{Ink} for interaction knowledge.
Nevertheless, the previous \oakinki{} has two major limitations:
\begin{enumerate*}[label=\arabic*)]
\item it lacks human demonstrations that cover the process of fulfilling those affordances, and 
\item it lacks complex manipulation tasks that involve multiple object affordances.
\end{enumerate*}

In this paper, we present \oakinkii{}, extending the data and methodology of the previous \oakinki{}.
In order to manage the inherent complexity in complex manipulation tasks,
\oakinkii{} adopts an \textit{object-centric} perspective and constructs three levels of abstraction upon manipulation tasks:
\begin{enumerate}[label=\textbf{\arabic*)}]
    \item \textbf{\textit{Affordance}}: object/object-part level functionalities that enable manipulation. For example, a bottle cap affords securing and unsecuring of the content in the bottle. 
    \item \textbf{\textit{Primitive Task (Primitive)}}: a ``minimal'' sequence of hand-object interaction that fulfills a given object's affordance. For instance, to fulfill the affordance: securing, one needs to either {\small\primitiveid{screw}} or {\small\primitiveid{press}} the cap onto the bottle's opening to form a seal that prevents leaking.
    \item \textbf{\textit{Complex Task}}: sequential combination of \textit{Primitives} to address the long-horizon and multi-goals manipulation tasks. 
    Tasks are characterized as ``complex'' for their goal requires more than one object affordance. 
    \textit{Complex Tasks} also detail the \textbf{dependencies} among the \textit{Primitives} and dictate the \textbf{order} in which they are executed. To illustrate, to pour the fluid from a sealed bottle, one must first {\small\primitiveid{unscrew}} the cap and then {\small\primitiveid{pour}} out the liquid.
\end{enumerate}

\noindent In this way, \oakinkii{} delineates \textit{Complex Tasks} as directed acyclic graphs, hereafter referred to as \textbf{\textit{Primitive} Dependency Graphs (PDG)}. Within these graphs, each node represents a \textit{Primitive}, serving to fulfill a specific affordance. The directed edges illustrate the sequence in which \textit{Primitives} must be executed to achieve task completion.

Build upon the above methodology, \oakinkii{} introduces a large-scale dataset for  bimanual object manipulation. It encompasses human demonstrations for complex task completion, with multi-view image streams and paired pose annotations for human body, hands and objects.
\oakinkii{} contains \totalseqnum{} sequences of real-world bimanual manipulation sequences, where \seqcnum{} of these sequences are for \textit{Complex Tasks}. 
These sequences contain \totalframenum{} frames from four different views (one egocentric and three allocentric views).
The dataset includes four manipulation scenarios, \totalobjnum{} objects and \totalsbjnum{} invited subjects in total.

The versatile and task-driven nature of \oakinkii{} enables a wide range of applications.
In this paper, we focus on the task and motion planning for Complex Task Completion (CTC).
CTC involves two notable components: 
\begin{enumerate*}[label=\textbf{\arabic*)}]
    \item text-based \textit{Complex Task} decomposition using \textit{Primitives} and 
    \item task-aware motion generation to fulfill each \textit{Primitive}.
\end{enumerate*}
For the first component, we design a task interpreter with Large Language Model (LLM) that can generate the PDG and program the execution order of these \textit{Primitives}, based on textual descriptions of the \textit{Complex Tasks}.
For the latter component, we propose a generalist Task-aware Motion Fulfillment model (TaMF) to generate the hand motion at \textit{Primitive} level, based on the task-related object trajectory.

In summary, our contributions are as follows:
\begin{itemize}
    \item We build an object-centric, three-level abstraction to structure and understand complex manipulation tasks, \ie \textit{Affordance}, \textit{Primitive} to fulfill affordance, and \textit{Complex Task} with \textit{Primitive} dependencies.
    \item We introduce \oakinkii{}, a large-scale real-world dataset for bimanual object manipulation with human demonstrations for both \textit{Primitives} and \textit{Complex Tasks}.
    \item We propose a task-oriented framework, CTC, for complex task and motion planning. CTC consists of a LLM-based task interpreter for \textit{Complex Task} decomposition and a diffusion-based motion generator for \textit{Primitive} fulfillment.
\end{itemize}

\section{Related Works}

\cheading{Hand-Object Interaction Datasets.}
The recent research community has witnessed the emergence of numerous datasets on hand-object interactions. 
Earlier datasets \cite{hasson2019obman,corona2020ganhand,brahmbhatt2020contactpose} focused on static hand-object interactions with limited diversity.
More recent datasets \cite{hampali2020ho3dv2,taheri2020grab,chao2021dexycb,kwon2021h2o,fan2023arctic, zhu2023contactart, liu2024taco} captured dynamic hand-object interactions, covering bimanual interactions \cite{kwon2021h2o,fan2023arctic} and interactions with articulated bodies \cite{fan2023arctic,zhu2023contactart}.
We pay particular attention to interaction datasets related to object affordances. 
\cite{corona2020ganhand} expressed affordances in grasp type labels. 
\cite{brahmbhatt2020contactpose,taheri2020grab,fan2023arctic} collected intention labels for interactions. 
\cite{yang2022oakink, jian2023affordpose} studied object affordance-based hand-object interaction and collected object segmentations and affordance labels. 
\cite{liu2024taco} studied hand-object interactions in tool-action-object pairs.
Our proposed \oakinkii{} captures both human demonstrations for minimal interaction fulfilling object affordance as \textit{Primitive}, and demonstrations for \textit{Complex Task} where these affordances are fulfilled in specific order constrained by their dependencies.

{\vspace{5pt}\noindent\textbf{Decomposition of Manipulation Tasks.}}
Decomposing complex manipulation tasks into multiple building blocks across different hierarchies represents a widely adopted paradigm in the research community. 
\cite{cheng2023league} utilize the symbolic interface of task planners to construct an abstract state space, facilitating the reuse of hierarchical skills.
\cite{xu2018neural,huang2019neural} decompose task specifications into hierarchical neural programs, which feature bottom-level programs as callable subroutines interacting with the environment. 
\cite{chen2023sequential} chain multiple dexterous policies for achieving long-horizon task goals. 
\cite{belkhale2024rt} adopt a language-based methodology for decomposing action hierarchies.
In our work, we introduce an object(affordance)-centric, three-level abstraction framework within \oakinkii{} for the decomposition of complex manipulation tasks into \textit{Primitives}.

{\vspace{5pt}\noindent\textbf{Motion Synthesis.}}
Motion synthesis involves obtaining credible and realistic human action sequences. 
There are plenty of works to generate human motions \cite{petrovich2021action,tevet2022human,petrovich2022temos}, even interactions \cite{taheri2022goal,wu2022saga,ghosh2023imos,taheri2024grip,li2023object,li2024favor} based on different probabilistic model backbones like cVAE or denoising diffusion.
In particular, \cite{taheri2024grip,li2023object,li2024favor} synthesize human motion based on the object motion, delegating the latter part to preceding models serving as inputs.
Inspired by these works, we propose a new task within \oakinkii{}: Task-aware Motion Fulfillment
This task requires the model to synthesize hand motion trajectories based on given textual task descriptions and object motions.

{\vspace{5pt}\noindent\textbf{Foundation Models in Manipulation Tasks.}}
Recent days we have seen a significant increase in the application of foundation models in completing manipulation tasks.
There are significant efforts for end-to-end foundation models \cite{brohan2022rt,rt22023arxiv,open_x_embodiment_rt_x_2023} that outputs control signals from visual and textual inputs.
Existing works \cite{huang2022language,brohan2023can,singh2023progprompt} also leverage the in-context learning and zero-shot generalization abilities of Large Language Models (LLMs) for action selection from an array of choices to realize an autoregressive achievement of planning.
Demonstration of LLM-based program generation for task completion in \cite{huang2023voxposer,singh2023progprompt,li2024favor} inspires us to explore the ability of LLMs to reason code for discerning interdependencies between object affordances in complex tasks, along with the sequence in which they are implemented.
Our \oakinkii{} introduce the decomposition of \textit{Complex Tasks} into interdependent affordance-based \textit{Primitives}, accompanied by their diverse image-textual descriptions.
Based on this, we show an application of \oakinkii{} in Complex Task Completion utilizing existing power of foundation models.

\begin{figure*}[!t]
    \centering
    \vspace{-0.0em}
    \begin{minipage}{1.00\linewidth}
        \centering
        \includegraphics[trim=000mm 000mm 000mm 000mm, clip=False, width=\linewidth]{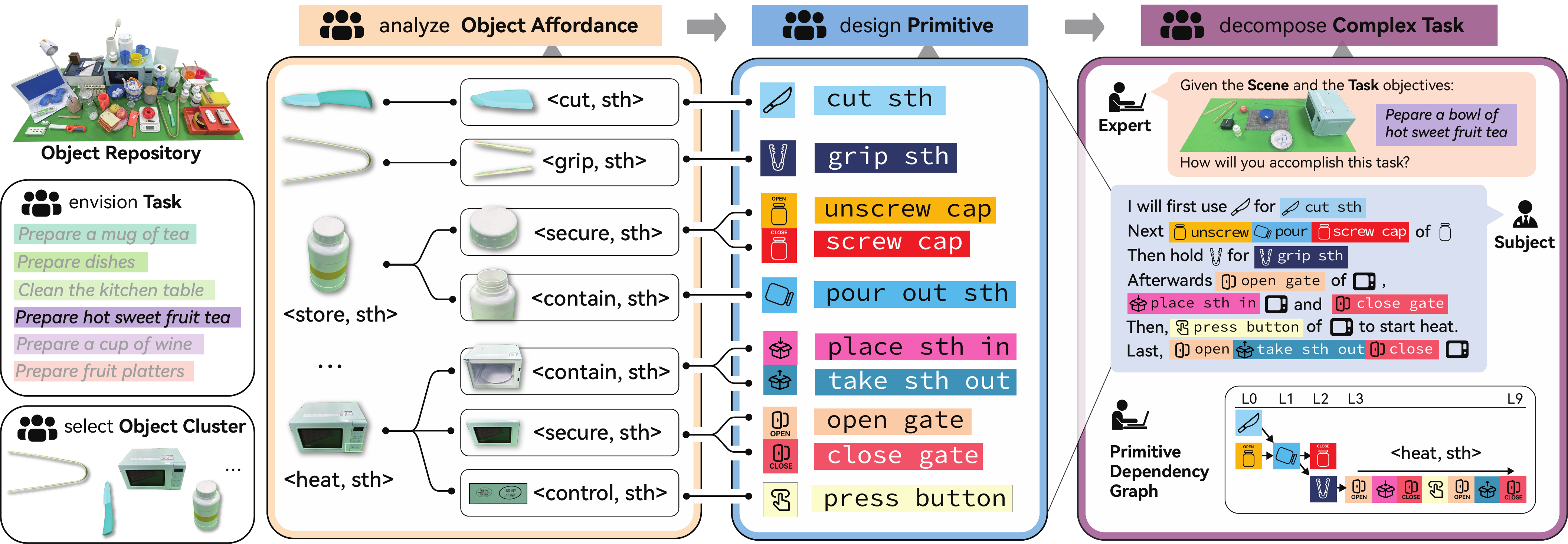}
    \end{minipage}
    \captionsetup{type=figure}
    \captionof{figure}{\textbf{Illustration of the complex task acquisition process.}
    This figure use a \textit{Complex Task}: `Prepare a bowl of hot sweet fruit tea.' to demonstrate the process. Initially, the annotators (\smash{\includegraphics[height=8pt]{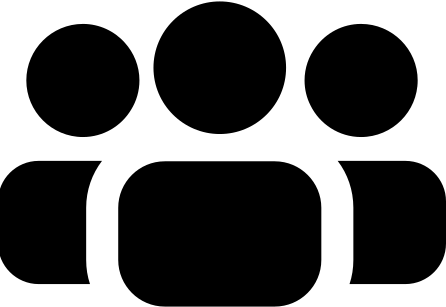}}) analyze the affordances of four essential objects (a gripper, a knife, a tea bottle, and a microwave oven) and design corresponding \textit{Primitive}. For instance, to prepare fruit slices, the \textit{Primitive}: \texttt{cut} associated with the knife blade is required. Following this, an expert (\smash{\includegraphics[height=8pt]{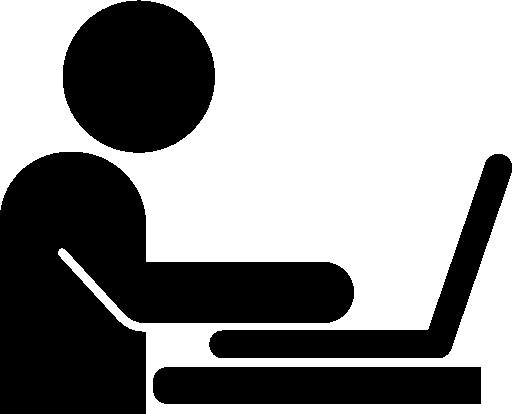}}) arranges the scene for the \textit{Complex Task}, and then the subject (\smash{\includegraphics[height=8pt]{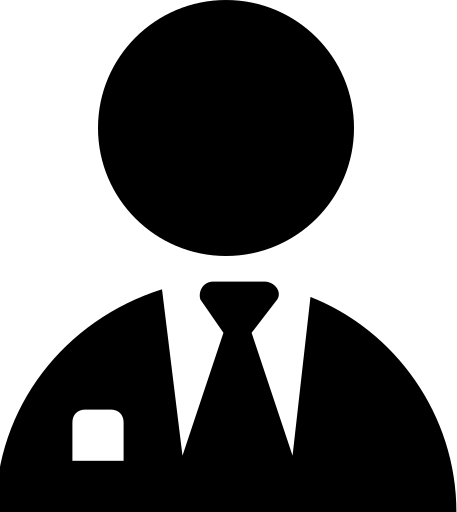}}), utilizing the designed \textit{Primitive}, plans the execution path of the \textit{Complex Task}. Later, these execution paths are structured into a \textit{Primitive} Dependency Graphs.}
    \label{fig:int_acq}
    \vspace{-0.5em}
\end{figure*}

\section{Construction of \oakinkii{}}
We first introduce how the three-level of abstractions are acquired in \cref{sec:int_acq}, then provide the details for data collection and annotation in \cref{sec:data_collection}. 
\subsection{Complex Task Acquisition}
\label{sec:int_acq}
\cheading{Task Initialization.}\label{sec:scene_construction}
Given a collected repository of objects, we first construct four manipulation scenarios. Each scenario has its unique characteristic and corresponds to a set of complex manipulation tasks. These scenarios are: 
\begin{enumerate*}[label=\arabic*)]
    \item kitchen table;
    \item study room table;
    \item demo chem lab;
    \item bathroom table.
\end{enumerate*}
Then, we invite four annotators (\smash{\includegraphics[height=8pt]{figure/intacq/annotators.png}}) to propose \textit{Complex Tasks} in these scenarios and select object cluster that required for these tasks (\cref{fig:int_acq}'s 1st column). 

\vspace{-0.5em}
\subsubsection{Object Affordance Analysis}
After the task targets are determined, we proceed to analyze the objects' affordances in given scenarios. The expression of affordance adheres to the definitions in the previous \oakinki{}~\cite{yang2022oakink}: each affordance contains a specific object part segmentation (\eg a bottle cap) and a descriptive phrase tuple (\eg \texttt{<secure, sth>}), which elucidates the function of that part. We provide examples of these affordances in \cref{fig:int_acq}'s 2nd column.

\vspace{-0.5em}
\subsubsection{Primitive Task Design} \label{sec:design_primitive_task}
In the second stage, We design \textit{Primitives} as the \textbf{minimal} interactions that fulfill those object affordance. 
Here ``minimal'' indicates the task are required to fully complete the functionality of a certain affordance without any redundant interaction process.
Each \textit{Primitive} contains a starting condition, a terminal condition, and the in-between hand-object interaction process.
For example, considering an affordance associated with a knife blade meant to \texttt{<cut, sth>}, 
a corresponding \textit{Primitive}, \primitiveid{cut}, requires the subject to move the blade to completely pass through the object to be cut so that the separated parts could be detached.
In this stage, we collect all available object affordances and their associated \textit{Primitives}, leading to a \textit{Primitive} tasks pool (\cref{fig:int_acq}'s 3rd column). 

\vspace{-1em}
\subsubsection{Complex Task Decomposition}\label{sec:decompose_combined_task}
In the third stage, we proceed to decompose the previous proposed \textit{Complex Task} -- characterized by its long-horizon and multi-goal manipulation targets -- into a series of 
short-term and single-goal \textit{Primitives}. 
In emphasizing the ordering of \textit{Primitive} completion is important for the \textit{Complex Task} completion, our approach also delineates the \textbf{dependencies} between \textit{Primitives}. 
Therefore, each  \textit{Complex Task} contains a series of \textit{Primitives}, along with a \textit{Primitive} Dependency Graph (PDG), which maps out the hierarchical execution order of these \textit{Primitives}. \textit{Primitives} at level 0 (L0) are independent, requiring no prior \textit{Primitives} to be completed, while the final level include those \textit{Primitives} that bring the \textit{Complex Task} to completion.

We deploy a dedicated protocol to acquire the decomposition and dependencies. As shown in \cref{fig:int_acq}'s 4th column, initially, an expert (\smash{\includegraphics[height=8pt]{figure/intacq/expert.png}}) instantiates the scene and target with specific description. Subsequently, a subject (\smash{\includegraphics[height=8pt]{figure/intacq/subject.png}}) is instructed to describe the order of the completion using the available \textit{Primitive} in the pool. Then, the expert records  and organizes this sequence into the PDG, concluding the \textit{Complex Task} acquisition process. 

\subsection{Data Collection and Annotation}
\label{sec:data_collection}
\begin{figure}[!t]
    \centering
    \vspace{-0.0em}
    \begin{minipage}{1.00\linewidth}
        \centering
        \includegraphics[trim=000mm 000mm 000mm 000mm, clip=False, width=\linewidth]{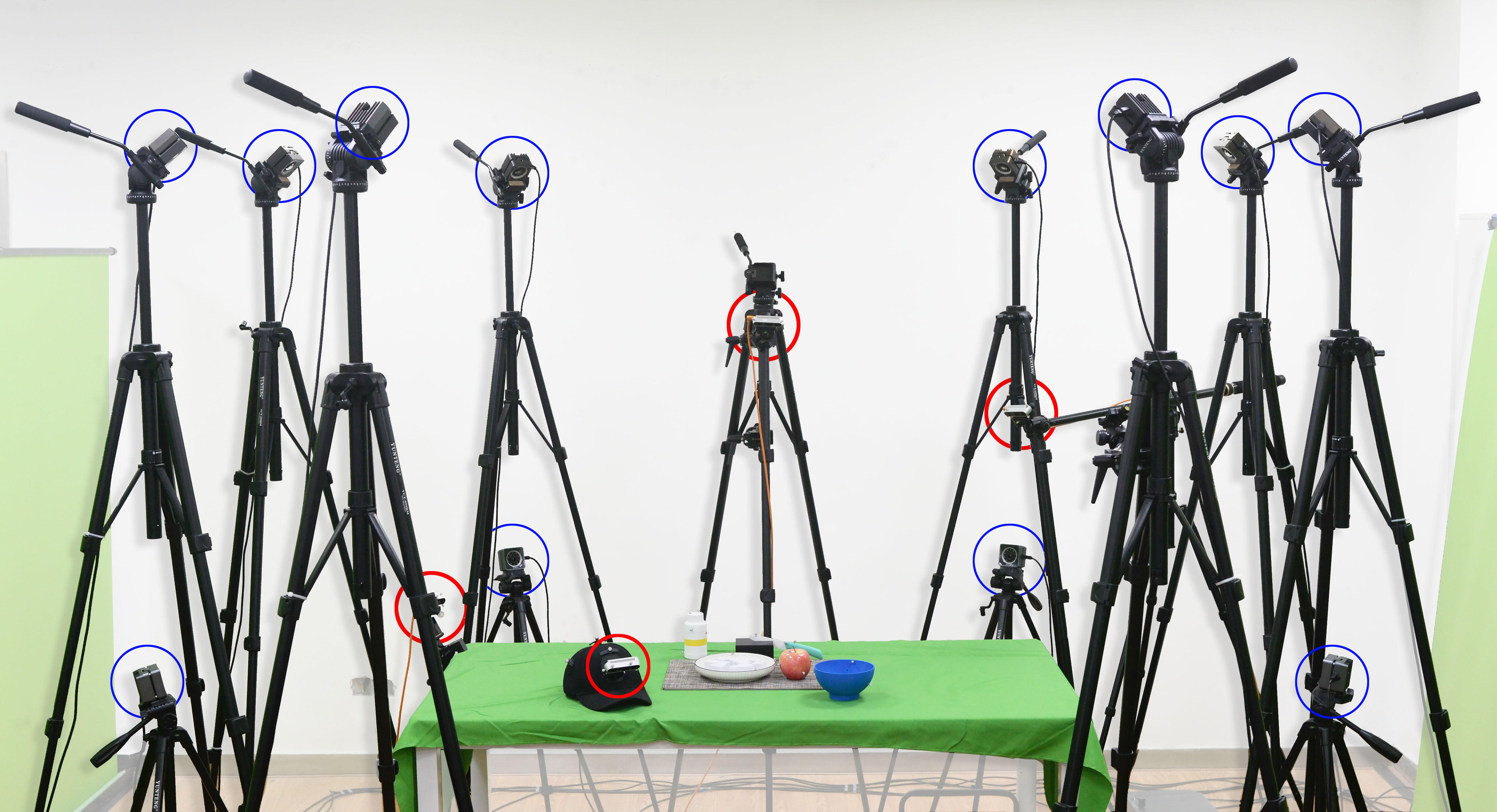}
    \end{minipage}
    \captionsetup{type=figure}
    \captionof{figure}{\cheading{Capture platform.} $12$ MoCap cameras are circled in blue and $4$ RGB cameras in red.}
    \vspace{-1.0em}
    \label{fig:capture_setup}
\end{figure}

After the acquisition of the three-level of abstractions, the subjects are required to complete the \textit{Primitive} and \textit{Complex Task} respectively in a data capture platform (\cref{fig:capture_setup}).
\vspace{-0.5em}\subsubsection{Capture Setup}\label{sec:cap_setup}
The data capture platform contains two major components: the multi-camera system for recording the manipulation process and the optical MoCap system for pose tracking. 
The MoCap system uses $12$ Optitrack Prime 13W infrared cameras to track the surface markers affixed to the subject's upper body, left and right hand, and interacting objects.
The multi-camera system consists of $4$ commodity RGB cameras, $3$ of which are from allocentric views and $1$ is from the egocentric view.
We synchronize all sensors at $\SI{30}{fps}$ and calibrate the transformation between these two systems.

\vspace{-0.5em}\subsubsection{Data Annotation}
\label{sec:anno_proc}
\cheading{Object Pose.} Poses of rigid bodies are directly solved via the MoCap system. 
For the poses of articulated bodies, the base parts of articulated bodies are handled similarly to rigid bodies, while the articulated parts are divided into two categories.
If the part is large enough to attach enough markers without blocking the interaction then it will be handled like rigid bodies.
Otherwise, only one marker is attached to that part. The marker's position is calibrated in the object's canonical coordinate frame. Later, given the articulation type (\eg revolution or prismatic), the parameter of the articulation joint is determined by minimizing the squared difference between the observed marker position and the recovered marker position in the object's canonical frame.

\qheading{Human Pose and Surface.} The annotation of human pose and surface relies on SMPL-X \cite{pavlakos2019expressive} body mesh. 
To actually acquire human pose and surface, we employ a two-stage fitting approach in align with the MoSH++ \cite{mahmood2019amass}.
In the first stage, we use the captured markers when the subject in T-pose to fit the subject's SMPL-X shape parameter $\bar{\beta}$ and each marker's location $P_{\mathcal{M}}^{(c)}$ in SMPL-X canonical space.
From stage one's optimization result, we can determine the correspondence $\mathcal{C} ( \cdot )$ from the subject's surface markers to the vertices of the SMPL-X model.
In the second stage, we fit the per-frame subject poses parameter $\theta$ throughout the task completion process. This fitting is grounded in the previously acquired shape $\bar{\beta}$ and marker correspondence $\mathcal{C} ( \cdot )$.
With pose and shape parameters obtained, the subject's body mesh is reconstructed using the SMPL-X model. Other body representations like MANO are derived from this result. Refer to \supp{} for details.

\begin{figure}[!t]
    \centering
    \vspace{-0.0em}
    \begin{minipage}{1.00\linewidth}
        \centering
        \includegraphics[trim=000mm 000mm 000mm 000mm, clip=False, width=\linewidth]{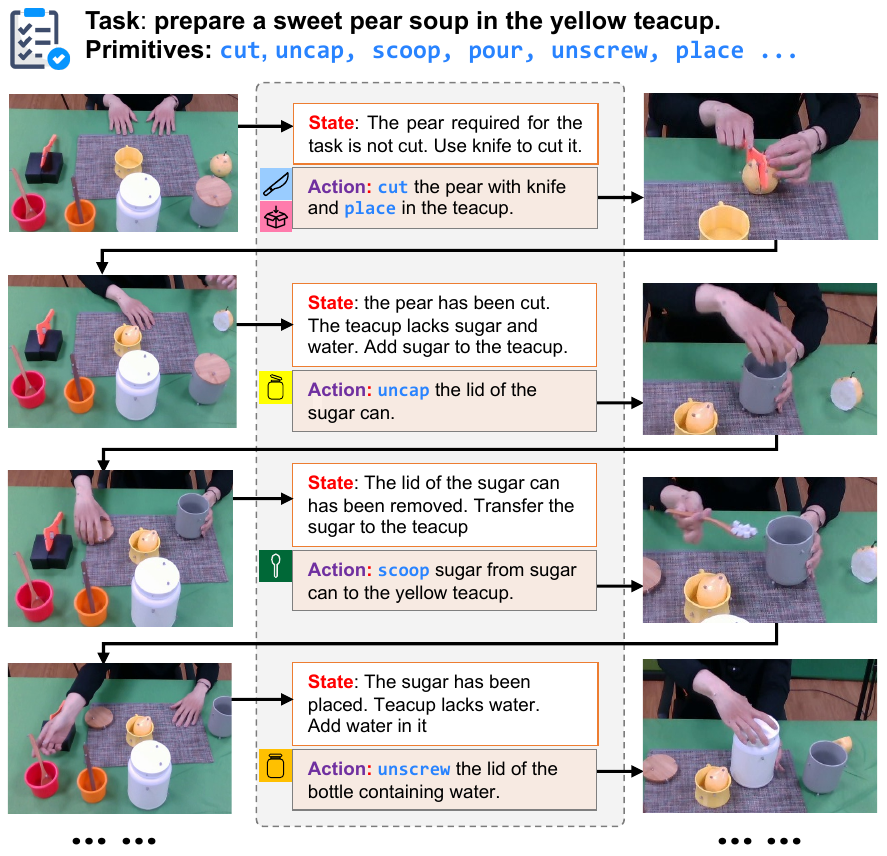}
    \end{minipage}
    \captionsetup{type=figure}
    \captionof{figure}{\cheading{Commentary of the task execution.} The left column shows the current state of the scene. The center column shows the narrative dialog retrieved from experts. The right column shows the upcoming \textit{Primitive} task to be executed.}
    \label{fig:task_narration}
    \vspace{-1.0em}
\end{figure}

\qheading{Commentary of Task Execution.}
\label{sec:narration_exec_path}
After the manipulation process is completed, we send the video recording to experts for analysis, requesting them to furnish detailed commentary on the task execution process. At each \textit{Primitive} step, experts are asked to provide comments on the current task state and the forthcoming action. 
Specifically, given the execution of the previous \textit{Primitive}, experts are asked to 1) summarize the tasks yet to be completed to achieve the manipulation goals, considering both the current scene and the upcoming \textit{Primitive} slated for execution; and 2) offer  descriptions of the next action using the available \textit{Primitives} in the pool.
This process is illustrated in \cref{fig:task_narration}.
The narrative text provided by experts are subsequently refined using GPT-4 \cite{openai2023gpt} to serve as commentary.
\oakinkii{} features on these commentaries as they encapsulate the expert's chain-of-thought when observing the manipulation process. These commentaries serve not only to interpret user behaviors but also to inform the generation of user actions. 

\begin{table*}[ht]
    \centering
    \footnotesize
    \setlength{\tabcolsep}{1.30pt}
    \resizebox{\textwidth}{!}{
        \begin{tabular}{l|cccccc|ccccccc|cc}
            \toprule
            \multirow{2}{*}{Dataset}                            &
            \multirow{2}{*}{\minitab[c]{image \\ mod.}}         &
            \multirow{2}{*}{resolution}                         &
            \multirow{2}{*}{\#frame}                            &
            \multirow{2}{*}{\#views}                            &
            \multirow{2}{*}{\#subj}                             &
            \multirow{2}{*}{\#obj}                              &
            \multirow{2}{*}{\minitab[c]{3D \\ gnd.}}            &
            \multirow{2}{*}{\minitab[c]{real / \\syn.}}         &
            \multirow{2}{*}{\minitab[c]{label \\method}}        &
            \multirow{2}{*}{\minitab[c]{hand \\pose}}           &
            \multirow{2}{*}{\minitab[c]{obj \\pose}}            &
            \multirow{2}{*}{\minitab[c]{afford. \\ inter.}}     &
            \multirow{2}{*}{\minitab[c]{dynamic \\ inter.}}     &
            \multirow{2}{*}{\minitab[c]{long- \\ horizon}}      &
            \multirow{2}{*}{\minitab[c]{task \\ decomp.}}       \\
                                                         &             &                   &                           &     &                &                &             &      &           &             &             &              &             &                           \\
            \midrule
            EGO4D~\cite{grauman2022ego4d}                & \greencheck & $\sim$                & $\sim$                       & 1   & 931            & --             & \redcross   & --   & --        & \redcross   & \redcross   & \redcross    & \redcross   & \greencheck & \greencheck \\
            HO3D~\cite{hampali2020ho3dv2}                & \greencheck & $640 \times 480$  & 78K                       & 1-5 & 10             & 10             & \greencheck & real & auto      & \greencheck & \greencheck & \redcross    & \greencheck & \redcross   & \redcross   \\
            GRAB~\cite{taheri2020grab}                   & \redcross   & --                & 1.62M                     & --  & 10             & 51             & \greencheck & real & mocap     & \greencheck & \greencheck & \yellowcheck & \greencheck & \redcross   & \redcross   \\
            H2O~\cite{kwon2021h2o}                       & \greencheck & $1280\times 720$  & 571K                      & 5   & 4              & 8              & \greencheck & real & auto      & \greencheck & \greencheck & \greencheck  & \greencheck & \redcross   & \redcross   \\
            HOI4D~\cite{liu2022hoi4d}                    & \greencheck & $1280\times 800$  & 3M                        & 1   & 9              & 1000           & \greencheck & real & crowd     & \greencheck & \greencheck & \greencheck  & \greencheck & \redcross & \redcross   \\
            ARCTIC~\cite{fan2023arctic}                  & \greencheck & $2800\times 2000$ & 2.1M                      & 9   & 10             & 11             & \greencheck & real & mocap     & \greencheck & \greencheck & \yellowcheck & \greencheck & \redcross   & \redcross   \\
            AssemblyHands~\cite{ohkawa2023assemblyhands} & \greencheck & $1920\times 1080$ & 3.03M                     & 12  & 34             & --             & \greencheck & real & semi-auto & \greencheck & \redcross   & \greencheck  & \greencheck & \greencheck & \greencheck \\
            Ego-Exo4D~\cite{grauman2023ego}              & \greencheck & $\sim$                & $\sim$                       & 5-6 & 839            & --             & \greencheck & real & semi-auto & \greencheck & \redcross   & \redcross    & \greencheck & \greencheck & \greencheck \\
            \oakinki-Image~\cite{yang2022oakink}         & \greencheck & $848 \times 480$  & 230K                      & 4   & 12             & {100}          & \greencheck & real & crowd     & \greencheck & \greencheck & \yellowcheck & \greencheck & \redcross   & \redcross   \\
            \midrule
            \textbf{\oakinkii{}}                         & \greencheck & $848 \times 480$  & \textbf{\totalframenum{}} & 4   & \totalsbjnum{} & \totalobjnum{} & \greencheck & real & mocap     & \greencheck & \greencheck & \greencheck  & \greencheck & \greencheck & \greencheck \\
            \toprule
        \end{tabular}
    }
    \caption{\small \textbf{A cross-comparison among various public datasets.} (Refer to \supp{} for the full table.)}
    \label{tab:dataset_comp}
\end{table*}
\section{The \oakinkii{} Dataset}
\vspace{-0.2em}
\subsection{Data and Annotation List}
\oakinkii{} provide RGB videos that record the manipulation processes. These videos are collected from multi-view (1 egocentric and 3 allocentric) setup, synchronized at $\SI{30}{fps}$, with resolution $848 \times 480$.
The annotations contains two parts: \textbf{1) 3D motion}, including pose and shape for the human upper-body, hands, and objects (with articulation parameters) during the interaction process; and the \textbf{2) task specification}, including object affordances, \textit{Primitives} that correspond to these affordances, \textit{Complex Tasks} with task goals, initial conditions, PDGs, expert commentary, and subject's completion sequence.
Evaluations of the 3D annotation qualities are provided in \supp. 
Annotation on 3D hand keypoints undergo cross-dataset validation with a reconstruction model, while the 3D poses associated with grasping actions are examined for the physical property integrity.

\subsection{Dataset Statistics}
\vspace{-0.2em}
\oakinkii{} sets up four scenarios of hand-object interaction with a total number of \targetnum{} long-horizon complex manipulation goals, which instantiates to \ctypenum{} \textit{Complex Tasks}.
\oakinkii{} contains in total \totalobjnum{} objects and \affordancenum{} affordance.
These affordances map to \ptypenum{} types of \textit{Primitives}.
\oakinkii{} contains \totalseqnum{} sequences of bimanual dexterous hand-object interaction in total.
\seqpnum{} of these are for \textit{Primitives} and \seqcnum{} are for \textit{Complex Tasks}.
In total, \oakinkii{} contains \totalframenum{} image frames.
We compare \oakinkii{} to multiple existing hand-object interaction datasets in \cref{tab:dataset_comp}. 
Here we highlight several notable features of \oakinkii{}:
1) it provides interaction grounded in object affordance (\vs HO3D, DexYCB);
2) it features long-horizon manipulation goals (\vs ARCTIC, HOI4D, GRAB);
3) it includes 3D pose and shape annotation for both hands and objects (\vs EGO4D, AssemblyHands); 
and 4) it offers task decomposition using \textit{Primitives}, which is not available in any datasets in \cref{tab:dataset_comp}.

\section{Selected Applications}

\subsection{Hand Mesh Reconstruction}

\label{sec:rec}
The Hand Mesh Reconstruction (HMR) task is to estimate the 3D hand pose during the interaction process from the captured images.
We benchmark HMR task under both single-view settings and multi-view settings.
In single-view settings, the image input only contains one view, egocentric or allocentric. 
In multi-view settings, the image input will contain multiple views, together with the camera calibration parameters.
For both settings we partition the corresponded task-specified subsets 
at the sequence level, maintaining the proportion of samples in train/val/test sets at approximately 70\%, 5\%, and 25\%.

We evaluate mean per joint position error (MPJPE), mean per vertex position error (MPVPE) in world space, wrist(root)-relative (RR) systems and systems after Procrustes analysis (PA).
We also evaluate area under curve (AUC) of correct keypoints percentage within range $0-\SI{20}{mm}$ in root-relative systems.
We show HMR benchmark results under both settings in \cref{tab:joint_hmr}.
\begin{table}[H]
    \centering
    \resizebox{\linewidth}{!}{
        \scriptsize
        \setlength{\tabcolsep}{1mm}
        \begin{tabular}{l|l|cccccc}
            \toprule
            \multirow{2}{*}{\textbf{Setting}} & \multirow{2}{*}{\textbf{Methods}}          & \textbf{PA-}          & \textbf{PA-}          & \textbf{RR-MPJPE}                 & \textbf{RR}            & \multirow{2}{*}{\textbf{MPJPE}} & \multirow{2}{*}{\textbf{MPVPE}} \\
                                              &                                            & \textbf{MPJPE}        & \textbf{MPVPE}        & (\textbf{AUC})                    & \textbf{-MPVPE}        &                                 &                                 \\
            \midrule
            \multirow{3}{*}{Mono}      & METRO~\cite{lin2021end}                    & 6.90                  & 6.47                  & 17.56    (0.410)                  & 16.44                  & --                              & --                              \\
                                              & RLE~\cite{li2021human}                     & \multirow{2}{*}{5.46} & \multirow{2}{*}{6.86} & \multirow{2}{*}{13.08    (0.441)} & \multirow{2}{*}{14.03} & \multirow{2}{*}{--}             & \multirow{2}{*}{--}             \\
                                              & \quad + HandTailer~\cite{lv2021handtailor} &                       &                       &                                   &                        &                                 &                                 \\
            \midrule
            \multirow{2}{*}{Multi}       & KP-based Fit~\cite{yang2023poem}     & 9.20                  & 8.83                  & 15.63 (0.349)                     & 15.38                  & 19.30                           & 19.11                           \\
                                              & POEM~\cite{yang2023poem}                   & 6.18                  & 6.61                  & 12.12 (0.581)                     & 12.15                  & 9.17                            & 9.52                            \\
            \bottomrule
        \end{tabular}
    }
    \vspace{0.0mm}
    \caption{\small Single- and multi-view HMR \textbf{evaluation results} in \textit{mm}.}
    \vspace{0.0mm}
    \label{tab:joint_hmr}
\end{table}

\subsection{Task-aware Motion Fulfillment (TaMF)}
\label{sec:tamf}
To achieve task objectives in interaction scenarios, we introduce a novel task: Task-aware Motion Fulfillment (TaMF). It targets at the generation of hand motion sequences that can fulfill given object trajectories conditioned on textual task descriptions.

\qheading{Task Formulation.} Given a textual description of the \textit{Primitive} task: $\mathbf{text}_{\mathrm{PT}}$, we assume the involved objects geometries $\mathbfcal{V}_o = \{\textbf{V}_{o,m}\}$ and their motion trajectories $\mathbfcal{T}_o = \{\textbf{T}_{o,m}^{(i)}\}$ during the interaction process are known.
We use subscript $h$ to represent human hands, $o$ to represent the object, $m$ to index different object instances (and different parts of the same instance) and superscript $(i)$ to index different timestamps.
The task is to generate a corresponding hands motion trajectory $\mathbfcal{P}_{h} = \{\textbf{P}_{h}^{(0:L)}\}$ conditioned on the textual description $\mathbf{text}_{\mathrm{PT}}$, object geometries $\mathbfcal{V}_o$, and motion trajectories $\mathbfcal{T}_o$.

\qheading{Evaluation Metrics.} We evaluate contact ratio (CR) and solid intersection volume (SIV) to measure the physical plausibility of the generated motion.
On sequence-level, we evaluate motion smoothness with Power Spectrum KL divergence of joints (PSKL-J) as in human motion generation, and evaluate FID to measure distances between the ground-truth motions and the generated motions. 
We also conduct a perceptual study to evaluate the level of realism for the generated motion.
Detailed definition of these metrics can be found in \supp.
\begin{figure}[!b]
    \centering
    \begin{minipage}{1.00\linewidth}
        \centering
        \includegraphics[trim=000mm 000mm 000mm 000mm, clip=False, width=\linewidth]{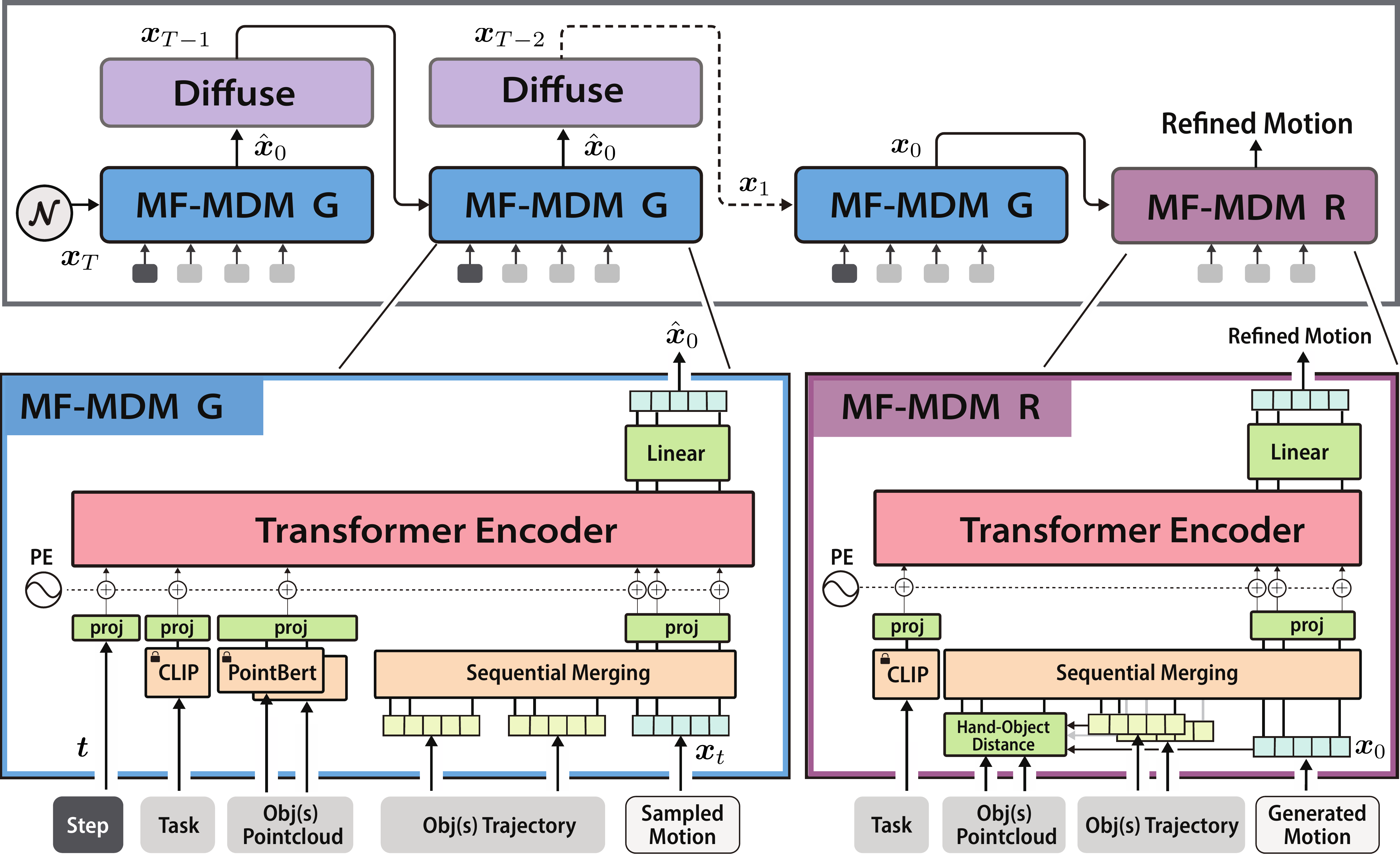}
    \end{minipage}
    \captionsetup{type=figure}
    \captionof{figure}{\small\textbf{Architecture of MF-MDM.} 
    First sample random noises $\boldsymbol{x}_T$; 
    then at each step iterating from $T$ to $1$, MF-MDM G predicts the cleaned sample $\hat{\boldsymbol{x}}_0$ and then diffuse it back to $\boldsymbol{x}_{t-1}$. 
    After the generated sample $\boldsymbol{x}_0$ is acquired, it is refined by MF-MDM R for better interaction details.
    }
    \label{fig:arch_mf_mdm}
\end{figure}

\begin{figure}[!t]
    \centering
    \vspace{-0.0em}
    \begin{minipage}{1.00\linewidth}
        \centering
        \includegraphics[trim=000mm 000mm 000mm 000mm, clip=False, width=\linewidth]{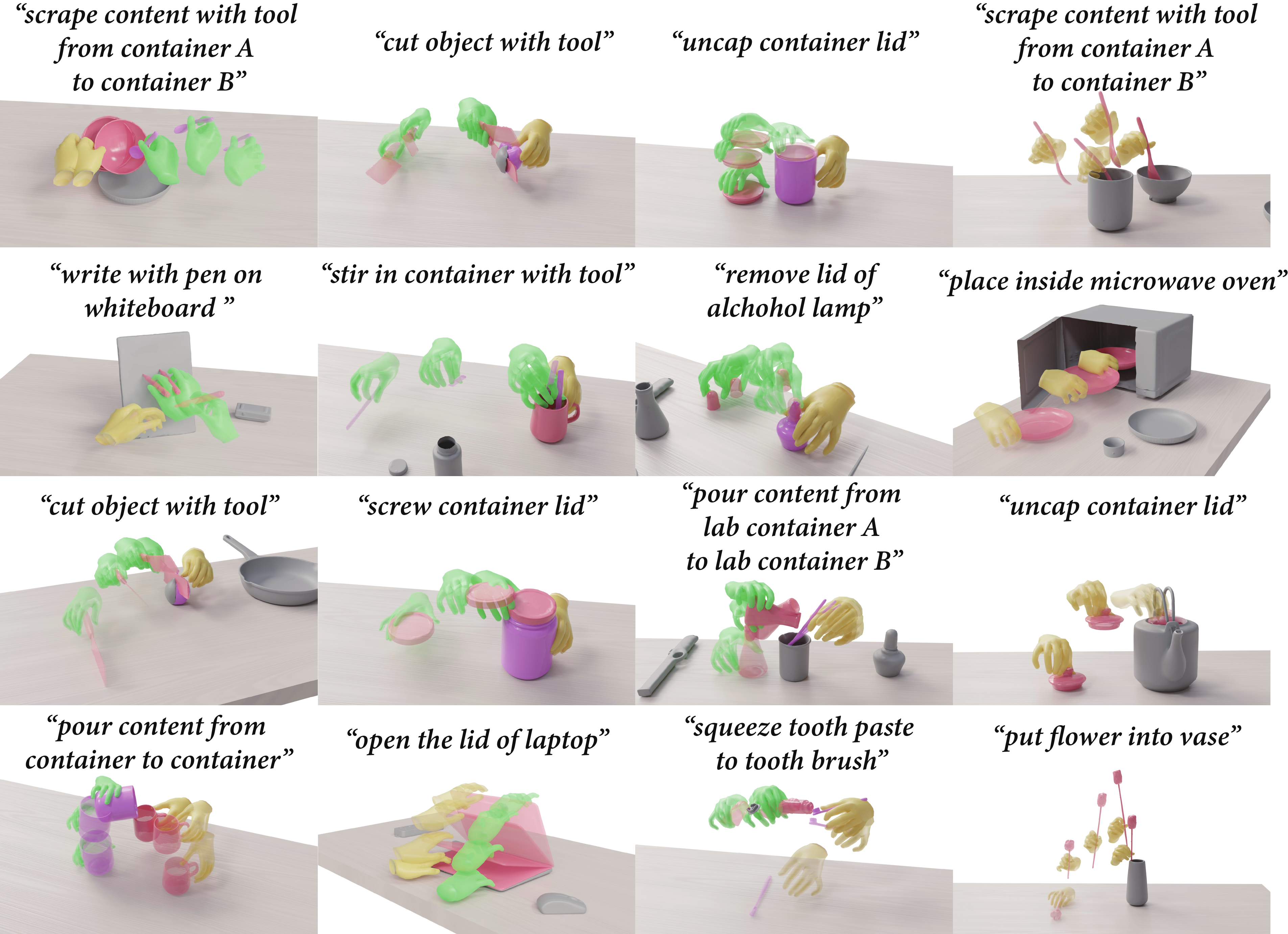}
    \end{minipage}
    \captionsetup{type=figure}
    \captionof{figure}{\textbf{Qualitative Visualization} of the generated hand motion in TaMF model.}
    \label{fig:tamf_qualitative}
\end{figure}

\begin{table}[!t]
    \centering
    \resizebox{0.8\linewidth}{!}{
        \makeatletter\def\@captype{table}\makeatother
        \setlength{\tabcolsep}{1mm}{
            \begin{tabular}{cc|cc}
                \toprule
                \multicolumn{2}{c|}{Physical lausibility} & \multicolumn{2}{c}{Motion Smoothness}                                                                            \\
                CR$\uparrow$                    & SIV (cm$^3$)$\downarrow$     & PSKL-J (\gt, \textbf{p.}) $\downarrow$ & (\textbf{p.}, \gt) $\downarrow$ \\
                \hline
                0.90                                      & 4.17                                  & 0.0446                                 & 0.0460                          \\
                \midrule
                \multicolumn{2}{c|}{\multirow{2}{*}{FID}} & \multicolumn{2}{c}{Perceptual Score}                                                                             \\
                                                          &                                       & Dataset                                & Generated                       \\
                \cline{1-4}
                \multicolumn{2}{c|}{1.369}                & $4.66 \pm 0.48$                       & $3.64 \pm 0.85$                                                          \\
                \bottomrule
            \end{tabular}
        }
    }
    \caption{\textbf{Evaluations} of generated hand motion in TaMF model.
        PSKL-J is evaluated between the training data (\gt) and the generated hand motion trajectory (\textbf{p.}); both directions are included as PSKL-J is an asymmetric metric.}
    \label{tab:tamf_quantitative}
\end{table}

\qheading{Model and Results.} We enhance a diffusion-based motion generation model: MDM \cite{tevet2022human}, tailoring it to the nuanced requirements of task-aware hand motion synthesis.
The model architecture is visualized in \cref{fig:arch_mf_mdm}. 
Our proposed model, named as MF-MDM, consists of two components: 
\begin{enumerate*}[label=\textbf{\arabic*)}]
    \item \textbf{MF-MDM G}, which generates human motion trajectory conditioned on textual descriptions of tasks and object motion trajectories; and
    \item \textbf{MF-MDM R}, which refines generated hand motion based on spatial hand-object relationships.
\end{enumerate*}
The sampling process is modeled as a reversed diffusion process of gradually cleaning noised samples.
The key difference for MF-MDM is to incorporate multi-object related probabilistic conditions into existing transformer encoder.
To achieve this, we employ an extra layer, Sequential Merging, to aggregate spatial relationships in the interaction scene at each frame.
The object motion trajectories and the previously diffused hand motion trajectory are projected to the same dimension and aggregated.
For the refine model MF-MDM R, we append hand-object distances as an extra spatial information for Sequential Merging layer.
The aggregated embedding sequence is combined with other tokens before being fed into the main transformer encoder: the noising step token, the text embedding of the task description from the CLIP text encoder, and the aggregated object geometry embeddings from the PointBert encoder.
We also provide the quantitative evaluations in \cref{tab:tamf_quantitative} and qualitative visualization in \cref{fig:tamf_qualitative}.

\begin{figure*}[t]
    \centering
    \vspace{-0.0em}
    \begin{minipage}{1.00\linewidth}
        \centering
        \includegraphics[trim=000mm 000mm 000mm 000mm, clip=False, width=\linewidth]{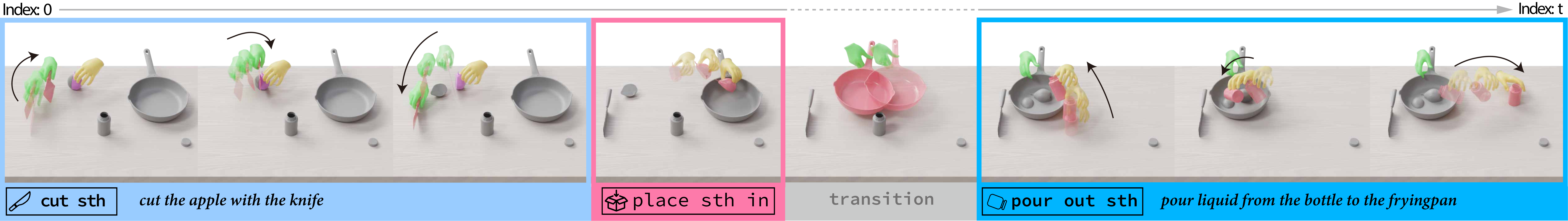}
    \end{minipage}
    \captionsetup{type=figure}
    \captionof{figure}{\cheading{Visualization of Motion Generation Outcome in Complex Task Completion.}
    }
    \label{fig:ctc_viz}
    \vspace{-0.5em}
\end{figure*}
\begin{figure}[!ht]
    \centering
    \vspace{-0.0em}
    \begin{minipage}{1.00\linewidth}
        \centering
        \includegraphics[trim=000mm 000mm 000mm 000mm, clip=False, width=\linewidth]{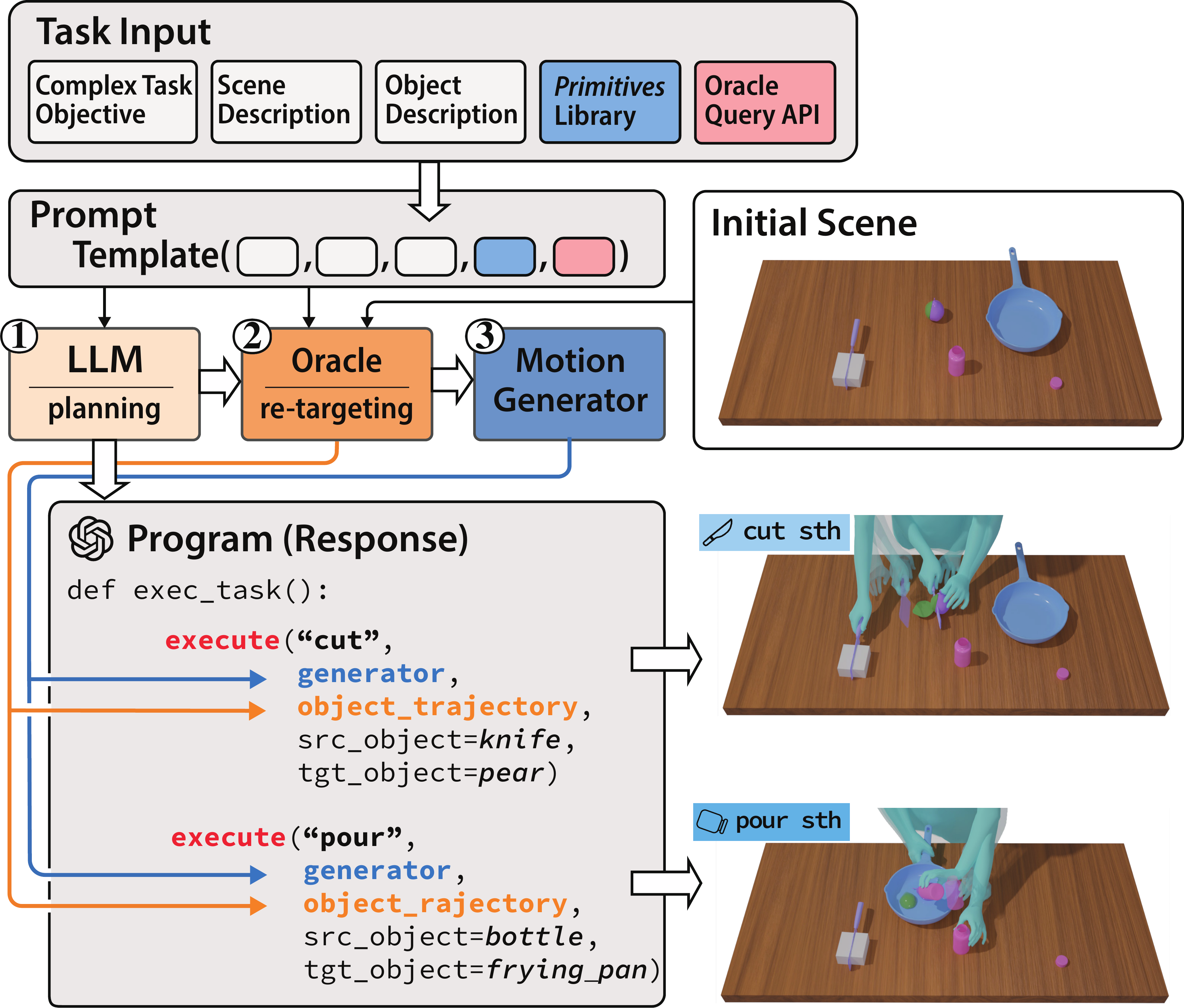}
    \end{minipage}
    \captionsetup{type=figure}
    \captionof{figure}{\cheading{The diagram of Complex Task Completion.} 
    The task input populates a predefined template to generate the prompt for planning.
    The $\customcircled{1}$ LLM (GPT-4) responds with code of the program's execution path, delineating the DAG for \textit{Primitive} dependency.
    Within the code response block, the \textcolor{objtraj}{\textbf{\texttt{orange}}} snippets marks the $\customcircled{2}$ Oracle to re-target object trajectories; the \textcolor{generator}{\textbf{\texttt{blue}}} snippets indicate $\customcircled{3}$ motion generators for \textit{Primitives}.
    }
    \vspace{-0.5em}
    \label{fig:ctc}
\end{figure}

\subsection{Complex Task Completion (CTC)}
\label{sec:ctc}
\oakinkii{} brings in a new application -- breaking \textit{Complex Task} goals into paths of \textit{Primitive} motions.
The Complex Task Completion (CTC) is to generate hands motion trajectories based on a textual description of the scene and the task objectives. Considering the challenge of direct translation from complex task and scene text to end-to-end motion generation, which involves a transition across multiple modalities, there is currently no adequate framework to address this problem. Therefore, we decompose CTC into three stages, tackling each one sequentially.

The process initially begins with text-based \textbf{1) \textit{Primitive} planning}.
The recent breakthroughs in foundation models \cite{openai2023gpt, zhen20243d}, such as Large Language Models (LLMs), allow us to utilize them as the task planner, as these models already have the capability to plan the \textit{Primitive} execution path, while only requiring proper guidance and context.
The output of this stage is a task planning script that includes the execution order for each \textit{Primitive}.
Subsequently, the problem is reformulated into generating the hand and object motion trajectories for each \textit{Primitive}, based on the target task and scene state, thus modeling $P(\mathbfcal{P}_h, \mathbfcal{T}_o | \mathbf{text}_{\mathrm{PT}})$. We again break this down into two subtasks: \textbf{2) object trajectory retrieval}, \ie $P(\mathbfcal{T}_o | \mathbf{text}_{\mathrm{PT}})$ and \textbf{3) hand motion generation} \ie $P(\mathbfcal{P}_h | \mathbfcal{T}_o, \mathbf{text}_{\mathrm{PT}})$. The former is solved by re-targeting\footnote{re-target refers to the process of adjusting pre-existing motion trajectories to align with new initial and target poses of objects, ensuring compatibility with the current scene} object motion from expert's demonstration to meet the newly generated random scene. The latter is our pre-defined Task-aware Motion Fulfillment model (TaMF, \cref{sec:tamf}). 

\qheading{$\customcircled{1}$ Primitive Planning by LLMs.}
In this stage, we leverage the off-the-shelf GPT-4 \cite{openai2023gpt} to generate program that decompose the \textit{Complex Task} as a sequence of \textit{Primitive}.
We first embed the scene description $\mathbf{text}_{\mathrm{scene}}$, the complex task description $\mathbf{text}_{\mathrm{goal}}$ and each object's description $\{\mathbf{text}_{\mathrm{obj}}\}$ into the prompt based on manually designed templates.
GPT-4 will respond to the prompt using the \textbf{program}. As shown in \cref{fig:ctc}'s code block, 
this program instantiates the \textit{Primitive} Dependency Graph (PDG) using a sequence of code snippets, where each node of the PDG (\textit{Primitive}) is implemented as a 
\inlinecode{\textbf{\textcolor{execute}{execute}}([primitive], ...)}function, and the edge of the PDG is implemented as function's calling order. 
Then we use a dependency checker built upon the PDG information in \oakinkii{} to test whether the generated program completes the \textit{Complex Task} without violation of constraints.
If a successful program is obtained, we move to the next stage.

At this moment, the \inlinecode{\textbf{\textcolor{execute}{execute()}}} function in \cref{fig:ctc}'s code snippets remain incomplete, lacking two pivotal components: the \inlinecode{\textbf{\textcolor{objtraj}{object\_trajectory}}} and the hand motion \inlinecode{\textbf{\textcolor{generator}{generator}}}. We will address these components in the following two stages.

\qheading{$\customcircled{2}$ Object Trajectories Retrieval from Oracle.}
Accomplishing a \textit{Primitive} task necessitates the object's motion trajectories within that context. In this stage, we leverage an Oracle to retrieve object motion trajectories based on a certain scene and \textit{Primitive}. The term ``Oracle'' denotes a dual-function capability: 
1) pursuant to a given \textit{Primitive}, it fetches the object motion trajectories within the \oakinkii{} dataset, and 2) it re-targets these expert-derived trajectories based on the initial, functional and post poses of the objects, thereby conforming to new scene requirements and generating the desired \inlinecode{\textbf{\textcolor{objtraj}{object\_trajectory}}}.

\qheading{$\customcircled{3}$ Hand Motion Generation with TaMF.}
Once the object trajectories are obtained, the final stage is to generate hand motion trajectories for each \textit{Primitive}. 
To this end, we utilize our previously designed Task-aware Motion Fulfillment model (TaMF, \cref{sec:tamf})  as a generalist  \inlinecode{\textbf{\textcolor{generator}{generator}}} (indicating that 
a singular TaMF model accommodates all \textit{Primitives}). 
After populating all \inlinecode{\textbf{\textcolor{execute}{execute()}}} functions with the determined object trajectories and generator, the program is executed in sequel and all the \textit{Primitive} trajectories are connected by interpolation. This interpolation ensures smooth transitions by linking the final state of a preceding trajectory with the initial state of the subsequent one. 

We show an example of the generated motions for \textit{Complex Task} in \cref{fig:ctc_viz}.
Details of test scene generation, prompts and templates, evaluations of primitive planning, success/failure cases are referred to \supp.

\section{Future Works}

\label{sec:limitation}

\oakinkii{} is a dataset packing a variety of hand-object interactions for human completion of long-horizon and multi-goal complex manipulation tasks.
\oakinkii{} incorporates \textit{Primitive} demonstrations, characterized as minimal interactions that fulfill object affordance, and \textit{Complex Tasks} demonstrations, which also include their decomposition into interdependent \textit{Primitives}.

First, we expect \oakinkii{} to support large-scale language-manipulation pre-training, improving the performance of multi-modal (\eg vision-language-action \cite{zhen20243d}) models for Complex Task Completion. In the longer term, we expect \oakinkii{} can potentially support learning frameworks capable of end-to-end text-to-manipulation generation.

Second, \oakinkii{} can empower various embodied manipulation tasks by re-targeting the collected demonstrations of \textit{Primitives} to different embodiments, such as heterogeneous hands and platforms as \cite{handa2020dexpilot, qin2021dexmv, xu2023unidexgrasp, wan2023unidexgrasp++, qin2022one} implied. The interaction scenarios constructed in \oakinkii{} can also be transferred and integrated into existing simulation environments \cite{Todorov2012MuJoCoAP,makoviychuk2021isaac} to support embodied learning on object manipulation.

\noindent\rule[0.3ex]{\linewidth}{1.0pt}

{\small\qheading{Acknowledgments.}  
This work was supported by the National Key Research and Development Project of China (No. 2022ZD0160102), National Key Research and Development Project of China (No. 2021ZD0110704), Shanghai Artificial Intelligence Laboratory, XPLORER PRIZE grants, 
and 2023 Shanghai Pujiang X Program Project (No. 23511103104).
}

\clearpage
{
    \small
    \bibliographystyle{ieeenat_fullname}
    \bibliography{main}
}

\clearpage
\renewcommand{\appendixpagename}{Supplementary Materials}
\begin{appendices}
\label{appendices}

\section*{Table of Contents}
\begin{enumerate}[font={\bfseries}, leftmargin=*]
    \item [\ref{sec:anno_detail}] Annotation Details
    \begin{enumerate}[font={\bfseries}, leftmargin=*]
        \item [\ref{sec:platform_calib}] Platform Calibration \& Synchronization
        \item [\ref{sec:data_cleaning}] Data Cleaning
        \item [\ref{sec:human_anno}] Human Pose and Surface
    \end{enumerate}
    \item [\ref{sec:data_statistics}] Dataset Meta Information
    \begin{enumerate}[font={\bfseries}, leftmargin=*]
        \item [\ref{sec:app_subsets}] Task-specific Subsets
    \end{enumerate}
    \item [\ref{sec:evaluation}] Dataset Evaluation
    \begin{enumerate}[font={\bfseries}, leftmargin=*]
        \item [\ref{sec:cross_validation}] Cross-Dataset Validation
        \item [\ref{sec:quality_assessment}] Physical Property Assessment
    \end{enumerate}
    \item [\ref{sec:app_benchmark}] Tasks and Benchmarks
    \begin{enumerate}[font={\bfseries}, leftmargin=*]
        \item [\ref{sec:app_tamf}] Task-aware Motion Fulfillment
    \end{enumerate}
    \item [\ref{sec:app_ctc}] Application: Complex Task Completion
    \item [\ref{sec:data_inspect}] Dataset Inspection
    \begin{enumerate}[font={\bfseries}, leftmargin=*]
        \item [\ref{sec:di_tasks}] Task List
        \item [\ref{sec:di_viz}] Visualization
    \end{enumerate}
\end{enumerate}

\section{Annotation Details}
\label{sec:anno_detail}

\subsection{Platform Calibration \& Synchronization}

\label{sec:platform_calib}

\begin{figure}[!bt]
    \centering
    \vspace{-0.0em}
    \includegraphics[trim=000mm 000mm 000mm 000mm, clip=False, width=0.8\linewidth]{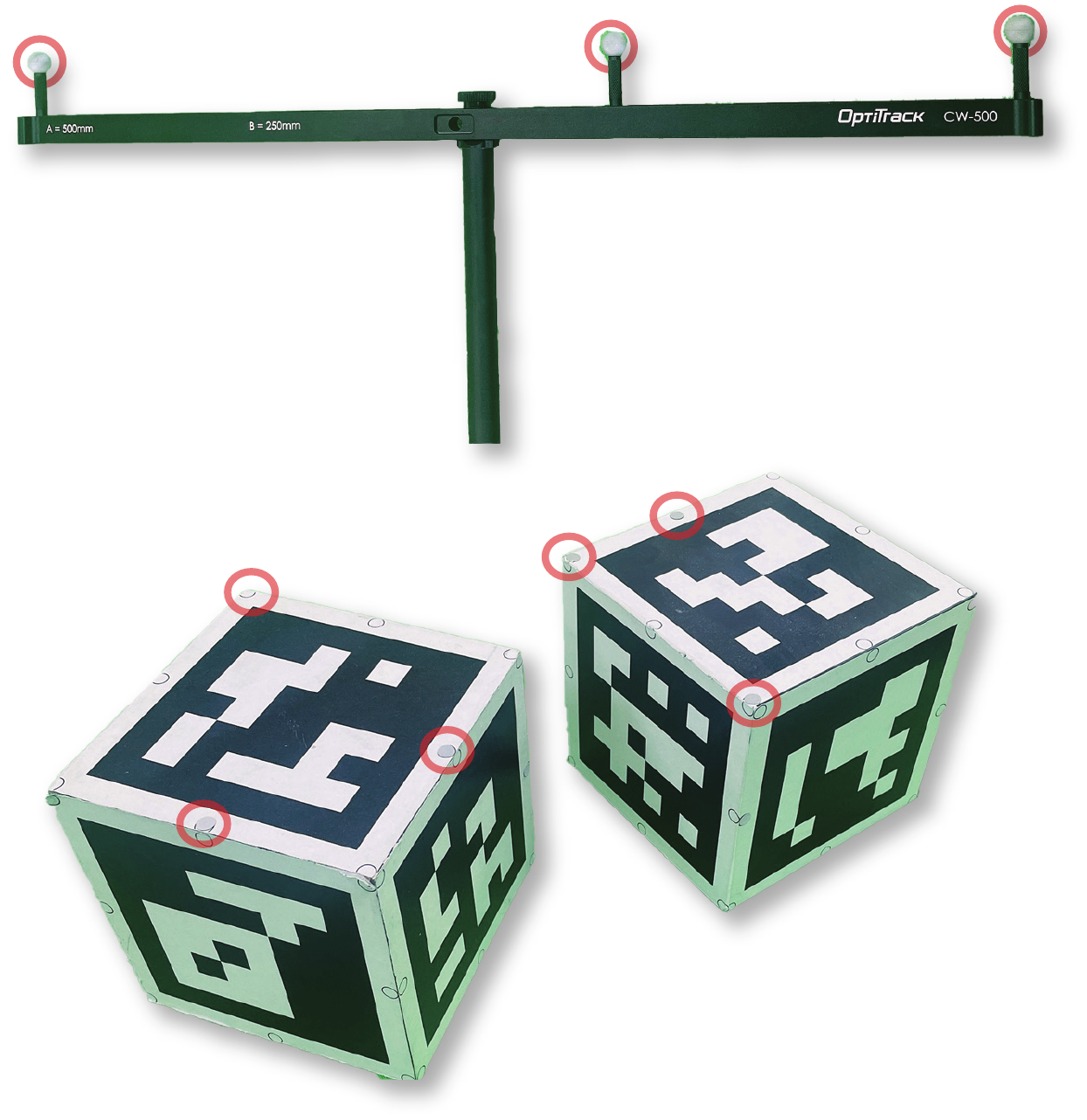}
    \captionsetup{type=figure}
    \captionof{figure}{\cheading{Illustration of Platform Calibration Tools.}
    The top is the vender-provided calibration wand for the MoCap system.
    The bottom are the ArUco cubes attached with reflective markers (circled in red).
    The ArUco patterns are for unifying the camera in the multi-camera system, while the surface-attached reflective markers are used for unifying the cameras with the MoCap system.
    }
    \vspace{-0.0em}
    \label{fig:calib}
\end{figure}

The MoCap system is calibrated via a specialized wand provided by the vendor. 
The cameras in the multi-camera system are calibrated using ArUco cubes.
These cameras are attached with reflective markers to be tracked by the MoCap system.
These calibration tools are shown in \cref{fig:calib}.
The two systems are time synchronized with software synchronization tools bundled in the ROS2 \cite{doi:10.1126/scirobotics.abm6074}.

\subsection{Data Cleaning}

\label{sec:data_cleaning}

In this section, we present a brief description of the process used to clean the reflective marker positions captured by the MoCap system, in preparation for subsequent object pose and human pose computations.
Inherent limitations of the MoCap system inevitably lead to errors in the reflective marker positions obtained: in extreme cases of occlusion, the system may fail to detect and record some markers; ghost markers may be included due to unwanted environmental reflections; when two or more markers come into close proximity, the system may incorrectly assign their labels or falsely identify them as a single marker.
These limitations lead to the introducing of a manual mechanism for cleaning and post-processing captured data.

The data cleaning procedure is composed of two components: \begin{enumerate*}[label=\arabic*)]
    \item the MoCap post-processing software;
    \item a multi-view interactive editor.
\end{enumerate*}
We invite three professional annotators for data cleaning.
The annotators first sequentially check the location of the captured reflective markers in the MoCap post-processing software \cite{motive}.
They then proceed to eliminate ghost points, split overlapping markers, correct mislabeled markers, and fill short gaps in the marker trajectories.
The annotators, following the order from articulated parts to rigid bodies, human bodies, and both hands, systematically clean the results of the collected markers.
Subsequently, the sequences are exported to the multi-view interactive editor.
In the editor, annotators verify the cleaned MoCap results and recover the marker positions in extreme occlusion cases through triangulation-based annotations from 2D point locations in multiple views.
The results are combined to get the cleaned captured reflective marker positions in the capture volume.

\subsection{Human Pose and Surface}

\label{sec:human_anno}

We employ a two-stage fitting approach inspired by the application of the MoSH++ algorithm in \cite{mahmood2019amass,taheri2020grab,fan2023arctic} for the SMPL-X \cite{pavlakos2019expressive} annotations.
The first stage registers the subject's SMPL-X \textit{shape} parameters and establishes correspondence mapping from the markerset to the surface of SMPL-X model.
The second stage registers SMPL-X \textit{pose} parameters for each frame in the sequence.
The two-stage fitting pipeline is implemented on PyTorch for its automatic differentiation support and common gradient descent based algorithms are used to solve for both stages.

\qheading{The first stage.} Let $\bar{\beta}$ be \textit{shape} parameters.
Let $P_{\mathcal{M}}^{(c)} \in \mathbb{R}^{N_{\mathcal{M}} \times 3}$ be surface marker positions lying in SMPL-X canonical space, where $N_{\mathcal{M}}$ is the number of markers in the target markerset.
Let $\theta=\{\theta_i\}$ be SMPL-X \textit{pose} parameters for each frame $i$ when the subject is in T-pose.
The first stage could be formulated as an optimization process to minimize the distance between the observed markers and the reconstructed markers derived from surface marker positions lying in SMPL-X canonical space, as shown in \cref{eqn:fit_stage1}.
\begin{equation}
    \label{eqn:fit_stage1}
    \begin{aligned}
        \min_{\theta, \beta, P_{\mathcal{M}}^{(c)}} E = \;
         & \lambda_1 E_{\text{recon}} \bigl( \theta, \beta, P_{\mathcal{M}}^{(c)} \bigr) + \lambda_2 E_{\text{prior(b)}} \bigl( \theta \bigr) + \\
         & \lambda_3  E_{\text{plau(h)}} \bigl( \theta \bigr) + \lambda_4 E_{\text{reg}} \bigl( \theta, \beta, P_{\mathcal{M}}^{(c)} \bigr)
    \end{aligned}
\end{equation}
The main cost is $E_{recon}$, which is the distance between the observed markers $P_\mathcal{M}$ and the reconstructed markers $\hat{P}_\mathcal{M}$ derived from surface marker positions.
Let $\boldsymbol{V}^{(c)}$ be the surface vertices in the canonical space of SMPL-X, $\boldsymbol{V}$ be the reconstructed surface vertices.
The markerset correspondence function $\mathcal{C}(\cdot)$ uses markerset position $P_{\mathcal{M}}^{(c)}$ and surface vertices $\boldsymbol{V}^{(c)}$ in canonical space to recover the markerset positions $\hat{P}_\mathcal{M}$ from the current reconstructed surface vertices $\boldsymbol{V}$.
It first projects the markers in canonical space $P_{\mathcal{M}}^{(c)}$ into local frames formed by the surface vertices to get the vertex index $I_{\mathcal{M}}$ and coefficients in local frames $C_{\mathcal{M}}$.
Then it uses the index to recover frames on posed vertices $\boldsymbol{V}$ and the coefficients in local frames to recover marker positions on the posed SMPL-X bodies.
$E_{recon}$ can then be expressed as in \cref{eqn:fit_stage1__recon}.
\begin{equation}
    \label{eqn:fit_stage1__recon}
    \begin{aligned}
        \boldsymbol{V}^{(c)} & = \text{SMPL-X} \big( \boldsymbol{0}, \bar{\beta} \big)
        \quad \quad
        \boldsymbol{V} =  \text{SMPL-X} \big( \theta, \bar{\beta} \big)                \\
        E_{\text{recon}}     & = \big\| P_\mathcal{M} - \hat{P}_\mathcal{M} \big\|^2
        =  \Big\| P_\mathcal{M} - \mathcal{C} \big(\boldsymbol{V},  P_{\mathcal{M}}^{(c)}; \boldsymbol{V}^{(c)} \big) \Big\|^2
    \end{aligned}
\end{equation}
$E_{\text{prior(b)}} \bigl( \theta \bigr)$, $E_{\text{plau(h)}} \bigl( \theta \bigr)$, and $E_{\text{reg}} \bigl( \theta, \beta, P_{\mathcal{M}}^{(c)} \bigr)$ are auxiliary cost terms.
$E_{\text{prior(b)}} \bigl( \theta \bigr)$ is an auxiliary term that minimizes negative log-likelihood of human body poses computed by prior from pre-existing datasets following the practice in \cite{mahmood2019amass}.
$E_{\text{plau(h)}} \bigl( \theta \bigr)$ is an implementation of anatomy loss in \cite{yang2021cpf} on the SMPLX model, designed to prevent distortion in the pose of the human body during the fitting process, enhancing its physical plausibility.
$E_{\text{reg}} \bigl( \theta, \beta, P_{\mathcal{M}}^{(c)} \bigr)$ is an auxiliary term that regularize the optimization variables.

\qheading{The second stage.} In the second stage, we fit the subject's pose $\theta = \{\theta_{t} \}$ throughout the interaction process based on the shape $\bar{\beta}$ and marker correspondence $\mathcal{C} ( \cdot )$ obtained in the first stage.

For each frame $t$ in the sequence, we optimize the subject's SMPL-X \textit{pose} parameter $\theta_{t}$ to minimize a combination cost composed of observed marker reconstruction error $E_{\text{recon}} \bigl( \theta_t \bigr)$, body pose prior $E_{\text{prior(b)}} \bigl( \theta_t \bigr)$, hand anatomy abnormality $E_{\text{plau(h)}} \bigl( \theta_t \bigr)$, hand-object intersection $E_{\text{plau(ho)}} \bigl( \theta_t \bigr)$ and other auxiliary regularization costs.
\begin{equation}
    \label{eqn:fit_stage2}
    \begin{split}
        \min_{\theta_t} E = \;
        & \lambda_1 E_{\text{recon}} \bigl( \theta_t \bigr) + \lambda_2 E_{\text{prior(b)}} \bigl( \theta_t \bigr) + \lambda_3 E_{\text{plau(h)}} \bigl( \theta_t \bigr) \\
        & + \lambda_4  E_{\text{plau(ho)}} \bigl( \theta_t \bigr) + \lambda_5 E_{\text{reg}} \bigl( \theta_t \bigr)
    \end{split}
\end{equation}
$E_{\text{plau(h)}} \bigl( \theta_t \bigr)$, $E_{\text{prior(b)}} \bigl( \theta_t \bigr)$, and $E_{\text{plau(h)}} \bigl( \theta_t \bigr)$ are the same cost terms as in the first stage. 
$E_{\text{plau(ho)}}$ penalizes the penetration and intersection between the interacting hands and objects by sampling internal points inside hand meshes and computing the sum of their signed distance function values of the objects.
$E_{\text{reg}} \bigl( \theta_t \bigr)$ not only includes regularization terms for the optimization variables but also contains velocity regularization terms to keep the smoothness of the annotated trajectories.

\qheading{Optimization} We implement the two-stage fitting pipeline on PyTorch for its automatic differentiation support.
We adopt Adam \cite{kingma2014adam} as the optimizer to solve for both stages, as it is widely applied and suitable for non-convex cost terms introduced in both stages.
We propose an early stopping mechanism for better running speed of the second stage: if there is no significant reduction in the fitting cost over a number of consecutive frames that exceeds a specific threshold, the optimization process will be terminated.

\begin{figure}[!bt]
    \centering
    \vspace{-0.0em}
    \begin{minipage}{1.00\linewidth}
        \centering
        \includegraphics[trim=035mm 035mm 035mm 015mm, clip=False, width=\linewidth]{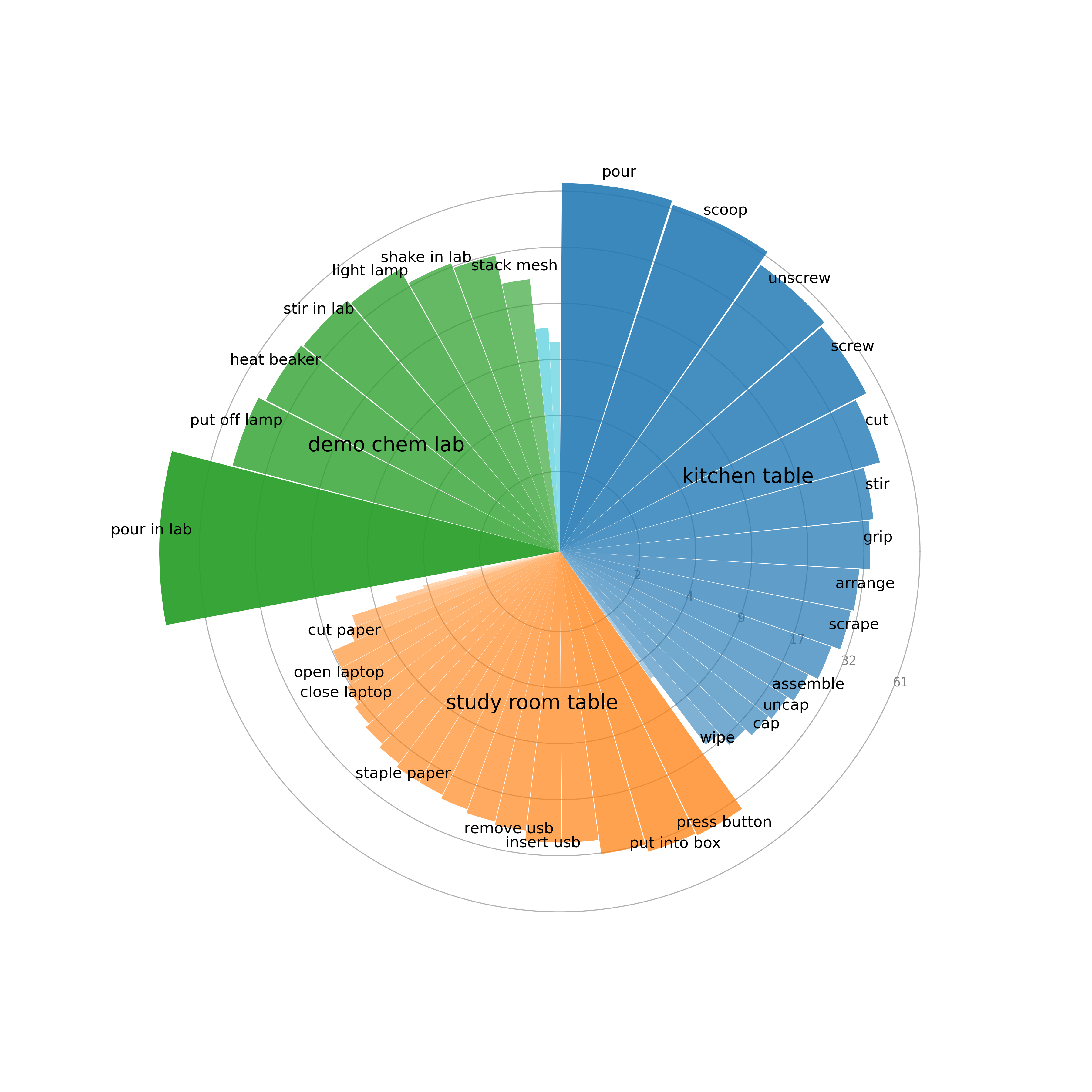}
        \includegraphics[trim=000mm 000mm 000mm 000mm, clip=False, width=\linewidth]{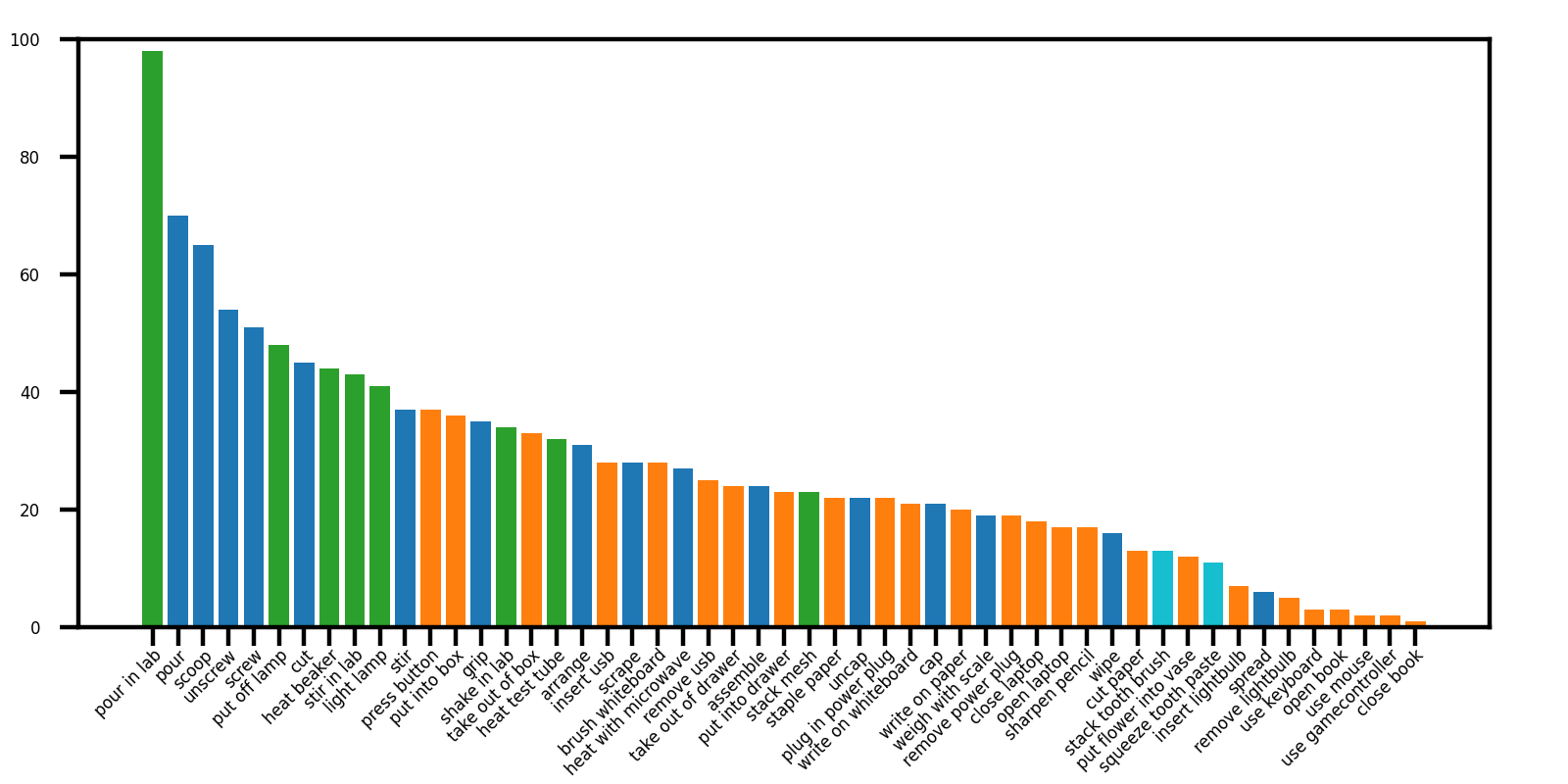}
    \end{minipage}
    \captionsetup{type=figure}
    \captionof{figure}{\cheading{Distribution of \textit{Primitive Task} demonstrations.} The sub-figure above displays the proportion of \textit{Primitive Task} demonstrations across various scenarios within the entire \oakinkii{} dataset, with frequently occurring \textit{Primitive Tasks} highlighted. The sub-figure below presents a list of \textit{Primitive Tasks} recorded in \oakinkii{}, along with the illustration of their corresponding quantity distribution. A list of all recorded \textit{Primitive Tasks} and \textit{Complex Tasks} can be found at \cref{tab:r_pt_ct} and \cref{tab:r_ct}.}
    \vspace{-0.0em}
    \label{fig:dist}
\end{figure}

\section{Dataset Meta Information}

\label{sec:data_statistics}

\begin{table*}[ht]
    \centering
    \footnotesize
    \setlength{\tabcolsep}{1.30pt}
    \resizebox{\textwidth}{!}{
        \begin{tabular}{l|cccccc|ccccccc|cc}
            \toprule
            \multirow{2}{*}{Dataset}                            &
            \multirow{2}{*}{\minitab[c]{image \\ mod.}}         &
            \multirow{2}{*}{resolution}                         &
            \multirow{2}{*}{\#frame}                            &
            \multirow{2}{*}{\#views}                            &
            \multirow{2}{*}{\#subj}                             &
            \multirow{2}{*}{\#obj}                              &
            \multirow{2}{*}{\minitab[c]{3D \\ gnd.}}            &
            \multirow{2}{*}{\minitab[c]{real / \\syn.}}         &
            \multirow{2}{*}{\minitab[c]{label \\method}}        &
            \multirow{2}{*}{\minitab[c]{hand \\pose}}           &
            \multirow{2}{*}{\minitab[c]{obj \\pose}}            &
            \multirow{2}{*}{\minitab[c]{afford. \\ inter.}}     &
            \multirow{2}{*}{\minitab[c]{dynamic \\ inter.}}     &
            \multirow{2}{*}{\minitab[c]{long- \\ horizon}}      &
            \multirow{2}{*}{\minitab[c]{task \\ decomp.}}       \\
                                                         &              &                              &                           &     &                &                &             &      &           &              &             &              &             &                           \\
            \midrule
            EPIC-KITCHEN-100~\cite{damen2022rescaling}   & \greencheck  & $\sim$                       & 20M$^\dagger$             & 1   & 37             & --             & \redcross   & --   & --        & \redcross    & \redcross   & \redcross    & \redcross   & \greencheck & \greencheck \\
            Ego4D~\cite{grauman2022ego4d}                & \greencheck  & $\sim$$^\ddagger$            & $\sim$$^\dagger$          & 1   & 931            & --             & \redcross   & --   & --        & \redcross    & \redcross   & \redcross    & \redcross   & \greencheck & \greencheck \\
            HA-ViD~\cite{zheng2023ha}                    & \greencheck  & $1280\times 720$             & 1.5M                      & 3   & 30             & 40             & \redcross   & --   & --        & \redcross    & \redcross   & \redcross    & \redcross   & \greencheck & \greencheck \\
                \midrule
            FPHAB~\cite{FirstPersonAction_CVPR2018}      & \greencheck  & $1920\times 1080$            & 105K                      & 1   & 6              & 4              & \greencheck & real & mocap     & \yellowcheck & \greencheck & \greencheck  & \greencheck & \redcross   & \redcross   \\
            ObMan~\cite{hasson2019obman}                 & \yellowcheck & $256 \times 256$             & 154K                      & 1   & 20             & 3K             & \greencheck & syn  & simulate  & \greencheck  & \greencheck & \redcross    & \redcross   & \redcross   & \redcross   \\
            YCBAfford~\cite{corona2020ganhand}           & \yellowcheck & --                           & 133K                      & 1   & 1              & 21             & \greencheck & syn  & manual    & \greencheck  & \redcross   & \redcross    & \redcross   & \redcross   & \redcross   \\
            HO3D~\cite{hampali2020ho3dv2}                & \greencheck  & $640 \times 480$             & 78K                       & 1-5 & 10             & 10             & \greencheck & real & auto      & \greencheck  & \greencheck & \redcross    & \greencheck & \redcross   & \redcross   \\
            ContactPose~\cite{brahmbhatt2020contactpose} & \greencheck  & $960 \times 540$             & 2.99M                     & 3   & 50             & 25             & \greencheck & real & auto      & \greencheck  & \greencheck & \yellowcheck & \redcross   & \redcross   & \redcross   \\
            GRAB~\cite{taheri2020grab}                   & \redcross    & --                           & 1.62M                     & --  & 10             & 51             & \greencheck & real & mocap     & \greencheck  & \greencheck & \yellowcheck & \greencheck & \redcross   & \redcross   \\
            DexYCB~\cite{chao2021dexycb}                 & \greencheck  & $640 \times 480$             & 582K                      & 8   & 10             & 20             & \greencheck & real & crowd     & \greencheck  & \greencheck & \redcross    & \greencheck & \redcross   & \redcross   \\
            H2O~\cite{kwon2021h2o}                       & \greencheck  & $1280\times 720$             & 571K                      & 5   & 4              & 8              & \greencheck & real & auto      & \greencheck  & \greencheck & \greencheck  & \greencheck & \redcross   & \redcross   \\
            HOI4D~\cite{liu2022hoi4d}                    & \greencheck  & $1280\times 800$             & 3M                        & 1   & 9              & 1000           & \greencheck & real & crowd     & \greencheck  & \greencheck & \greencheck  & \greencheck & \redcross   & \redcross   \\
            ARCTIC~\cite{fan2023arctic}                  & \greencheck  & $2800\times 2000$            & 2.1M                      & 9   & 10             & 11             & \greencheck & real & mocap     & \greencheck  & \greencheck & \yellowcheck & \greencheck & \redcross   & \redcross   \\
            ContactArt~\cite{zhu2023contactart}          & \yellowcheck & --                           & 332K                      & --  & --             & 80             & \greencheck & real & transfer  & \greencheck  & \greencheck & \redcross    & \greencheck & \redcross   & \redcross   \\
            AssemblyHands~\cite{ohkawa2023assemblyhands} & \greencheck  & $1920\times 1080$            & 3.03M                     & 12  & 34             & --             & \greencheck & real & semi-auto & \greencheck  & \redcross   & \greencheck  & \greencheck & \greencheck & \greencheck \\
            AffordPose~\cite{jian2023affordpose}         & \redcross    & --                           & --                        & --  & --             & 641            & \greencheck & syn  & manual    & \greencheck  & \greencheck & \greencheck  & \redcross   & \redcross   & \redcross   \\
            TACO~\cite{liu2024taco}                      & \greencheck  & $4096\times 3000$$^\ddagger$ & 5.2M                      & 13  & 14             & 196            & \greencheck & real & auto      & \greencheck  & \greencheck & \greencheck  & \greencheck & \redcross   & \redcross   \\
            Ego-Exo4D~\cite{grauman2023ego}              & \greencheck  & $\sim$$^\ddagger$            & $\sim$$^\dagger$          & 5-6 & 839            & --             & \greencheck & real & semi-auto & \greencheck  & \redcross   & \redcross    & \greencheck & \greencheck & \greencheck \\
            \oakinki-Image~\cite{yang2022oakink}         & \greencheck  & $848 \times 480$             & 230K                      & 4   & 12             & {100}          & \greencheck & real & crowd     & \greencheck  & \greencheck & \yellowcheck & \greencheck & \redcross   & \redcross   \\
            \oakinki-Shape~\cite{yang2022oakink}         & \redcross    & --                           & --                        & --  & --             & 1700           & \greencheck & real & transfer  & \greencheck  & \greencheck & \yellowcheck & \redcross   & \redcross   & \redcross   \\
                \midrule
            \textbf{\oakinkii{}}                         & \greencheck  & $848 \times 480$             & \textbf{\totalframenum{}} & 4   & \totalsbjnum{} & \totalobjnum{} & \greencheck & real & mocap     & \greencheck  & \greencheck & \greencheck  & \greencheck & \greencheck & \greencheck \\
            \toprule
        \end{tabular}
    }
    \caption{\small \textbf{A cross-comparison among various public datasets.}
        $\sim$: The value is either not provided on the paper or measured in a different unit.
        $\dagger$: Datasets measure in record time rather than number of captured frames.
        In particular, EPIC-KITCHEN-100 contains more than 100 hours of video, Ego4D 3670 hours, and Ego-Exo4D 1422 hours.
        They are larger in scale than any other dataset listed in the table.
        $\ddagger$: Dataset has a mixed resolution.
        \\
        \textbf{Legend}: \\
        \cheading{image mod.}: Image Modality. \greencheck means real image captures; \yellowcheck means synthetic (rendered) images; \redcross means no image modality provided. \\
        \cheading{3D gnd.}: 3D grounding. \greencheck means the dataset contains 3D grounding annotations; \redcross means the dataset is 2D only. \\
        \cheading{real / syn.}: Interaction is real / synthetic. Here syn indicates the interactions come from certain grasp/interaction synthesizer. \\
        \cheading{label method}: Label Method of 3D grounding information.
        For synthetic interactions, ``simulate'' indicates interactions are retrieved from physical-based grasp simulators, \eg GraspIt! \cite{miller2004graspit}; ``manual'' indicates interactions are labeled with human labor.
        For real interactions, ``mocap'' indicates the interactions are captured by MoCap systems; ``crowd'' indicates the interactions are derived from crowd-source keypoint annotations; ``auto'' indicates the interactions are retrieved from automatic annotation pipelines; ``semi-auto'' indicates a hybrid of ``crowd'' and ``auto'' methods. \\
        \cheading{afford. inter.}: Affordace-based Interaction. \greencheck means the interactions captured are affordance-aware and explicitly labeled; \yellowcheck means the interactions are afforance-aware but grouped in coarse-grained labels like intentions; \redcross means the interactions are not organized by object afforances. \\
        \cheading{dynamic inter.}: Dynamic Interaction. \greencheck means the dataset captures dynamic sequence of hand-object interactions; \redcross means the dataset captures static grasps that do not change during the interaction process. \\
        \cheading{long-horizon}: Long-horizon Tasks. As in the main text, \greencheck means the dataset contains captured interactions that involved more than one object afforances; \redcross vice versa. \\
        \cheading{task decomp.}: Task Decomposition. As in the main text, \greencheck means the dataset contains annotations that decomposition a complex task into multiple segments; \redcross vice versa.
    }
    \label{tab:dataset_comp}
\end{table*}

\subsection{Task-specific Subsets}

\label{sec:app_subsets}

Since \oakinkii{} is intended for various types of tasks, we create multiple subsets with different strategies for sample selection and data organization tailored to each specific task.
To obtain these subsets, we apply a few heuristics to determine whether each sample meets the requirements of the task it needs to support.
For instance, the visibility of hands or objects in image samples is essential for supporting related vision tasks.
We verify the individual and combined segmentation masks for hands and objects in the images.
If the proportion of the combined segmentation mask to its individual counterpart exceeds a certain threshold, we consider the instance as \textit{visible} in the current frame. 
We regard the object to interact as \textit{grasped} if it is close enough to hands {(minimal distance \footnotesize{$\le \SI{5}{mm}$})} and \textit{lifted} {(height displacement to the initial state \footnotesize{$\ge \SI{5}{mm}$})}.

We show the features and the construction methods for task-specific subsets in the following list. 

\qheading{\oakinkii{}-H-SV} Subset for hand reconstruction from single-view images. We select views that the subjects' hands are \textit{visible} to form this subset.
This subset supports task single-view Hand Mesh Recovery.

\qheading{\oakinkii{}-H-MV} Subset for hand reconstruction from multi-view images. We select combined views from different cameras as a single sample in this subset if the subjects' hands are \textit{visible} in a majority of camera views.
This subset supports task multi-view Hand Mesh Recovery.

\qheading{\oakinkii{}-HO} Subset for hand-object pose estimation or reconstruction from images. We select views that the subjects' hands are \textit{visible} and the object is \textit{grasped} to form this subset.

\qheading{\oakinkii{}-Grasp} Subset for grasps on the objects. We select frames that the object is \textit{grasped} to form this subset.

\qheading{\oakinkii{}-Motion-Approach\&Retreat} Subset for the interaction process that the subjects approach and grab the object for future tasks. We select frames from the sequence in one \textit{Primitive Task} that cover the process of \textit{approach} and \textit{grasp} the object are collected to form this subset.
This subset provides auxiliary information in Task-aware Motion Fulfillment and Object Trajectories Retrieval from Oracle Queries in Complex Task Completion.

\qheading{\oakinkii{}-Motion-Task} Subset for the interaction process that the subjects complete a task and fulfill one object affordance. We select frames from the sequence in one \textit{Primitive Task} that cover the process from the \textit{grasp} of the object to the \textit{completion} of the task are collected to form this subset.
This subset supports Task-aware Motion-Fulfillment.

\section{Dataset Evaluation}
\label{sec:evaluation}

Annotation on 3D hand keypoints undergo cross-dataset validation with a reconstruction model, while the 3D poses associated with grasping actions are examined for their physical property integrity.

\subsection{Cross-Dataset Validation}

\label{sec:cross_validation}

We perform cross-dataset validation to verify the consistency of 3D hand keypoint annotations in \oakinkii{} with pre-existing datasets.
We train a single-view hand mesh recovery model \cite{lin2021end} separately on three different training schemes: FreiHAND \cite{zimmermann2019freihand} only, \oakinkii{}-H-SV only, and a mixture of these two sets. 
We evaluate MPJPE and MPVPE after Procrustes analysis on \oakinki{}-image (\texttt{SP2}) \cite{yang2022oakink}, and the results shown in \cref{tab:cross_dataset_validation} indicates a consistent improvement on these metrics, verifying that \oakinkii{} complements existing datasets and boosts existing models.

\begin{table}[H]
    \renewcommand{\arraystretch}{1.0}
    \centering
    \small
    \resizebox{\linewidth}{!}{
        \setlength{\tabcolsep}{4.3pt}{
            \makeatletter\def\@captype{table}\makeatother
            \setlength{\tabcolsep}{1mm}{
                \begin{tabular}{l|c|cc}
                    \toprule
                    \textbf{Train}                              & \textbf{Test}                                     & {\footnotesize \textbf{PA-MPJPE} (\textit{mm}) $\downarrow$ } & {\footnotesize \textbf{PA-MPVPE} (\textit{mm}) $\downarrow$ } \\
                    \midrule
                    1) FreiHAND                                     & \oakinki{}-image (\texttt{SP2})                   & 12.07                                                         & 11.96                                                         \\

                     2) \oakinkii{}-H-SV        &  \oakinki{}-image (\texttt{SP2})  & 12.60                                         & 11.04                                         \\

                     \textbf{1) \& 2) mixture} &  \oakinki{}-image (\texttt{SP2}) & \textbf{10.94}                               & \textbf{9.67}                                \\
                    \bottomrule
                \end{tabular}
            }
        }}
    \vspace{0 mm}
    \caption{\textbf{Cross dataset validation} for \oakinkii{}.}
    \vspace{0 mm}
    \label{tab:cross_dataset_validation}
\end{table}

\subsection{Physical Property Assessment}

\label{sec:quality_assessment}

\begin{figure*}[t]
    \centering
    \begin{minipage}{1.00\linewidth}
        \centering
        \includegraphics[trim=000mm 000mm 000mm 000mm, clip=False, width=\linewidth]{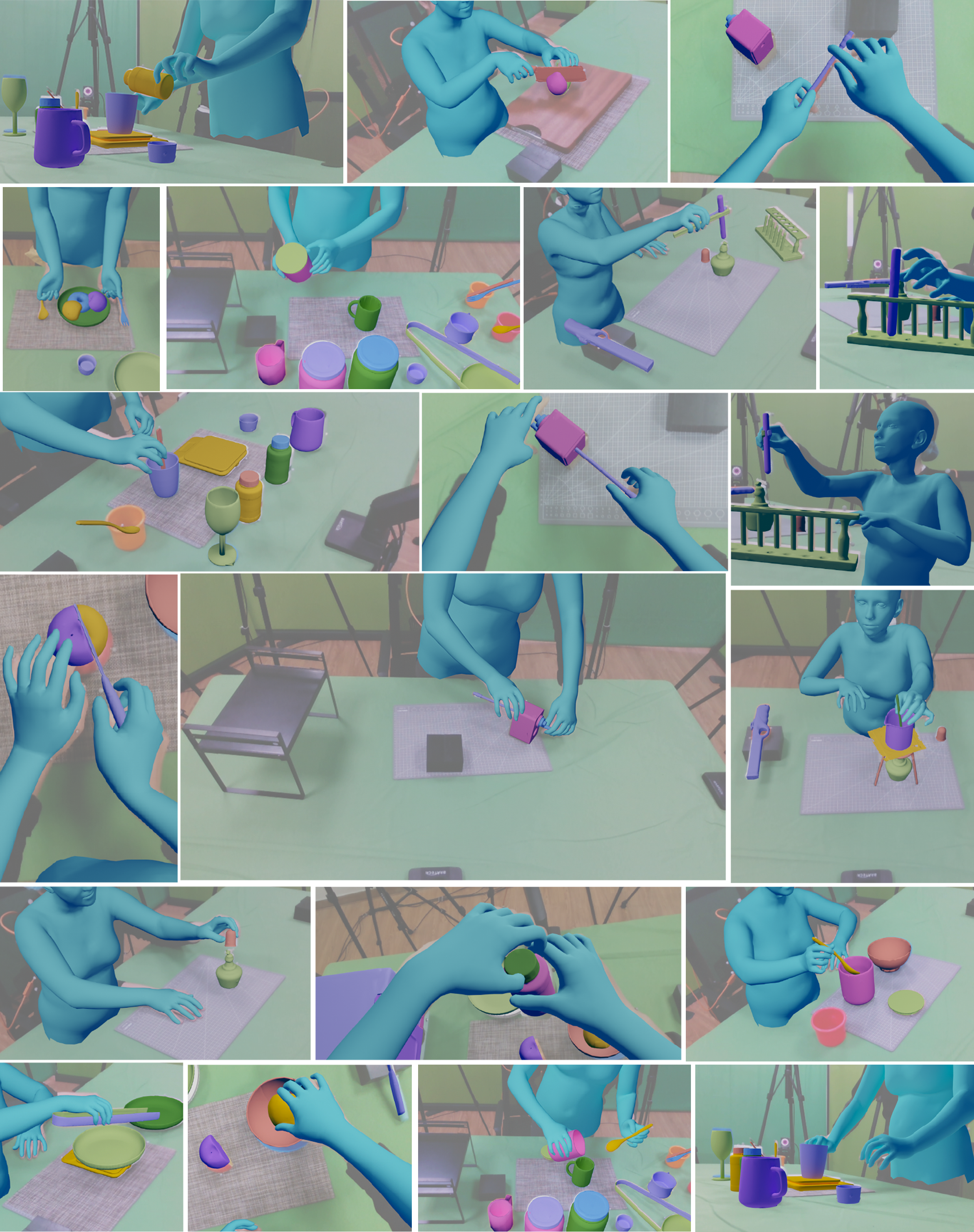}
    \end{minipage}
    \captionsetup{type=figure}
    \captionof{figure}{\cheading{Dataset visualization.} Human bodies and objects within the scene are rendered onto the captured raw images for visualization.}
    \vspace{0.0em}
    \label{fig:data_quality_viz2}
\end{figure*}

To evaluate the quality of the 3D poses associated with grasping actions in \oakinkii{}, we inspect several physical-based metrics that assess the feasibility and stability of captured hand-object interactions.
We restrict the samples to be evaluated based on certain rules (the objects in interaction need to be grasped and lifted), ensuring that these physics-based quality metrics accurately reflect the quality of the dataset during the interaction process. 
We compare \oakinkii{}-Grasp (-G.) with two subsets of \oakinki{}: \oakinki{}-Core and \oakinki{}-Shape (\cref{tab:dataset_quality}). 
We observe that, despite the use of the mocap system as the primary annotation method for easily scaling up the capture process, \oakinkii{} still achieved annotation quality on par with \oakinki{} built upon the hybrid of manual and mocap annotation.
More qualitative visualizations of \oakinkii{} are provided in \cref{fig:data_quality_viz2}.

\begin{table}[h]
    \centering
    \scriptsize
    \resizebox{\linewidth}{!}{
        \setlength{\tabcolsep}{4.3pt}{
            \makeatletter\def\@captype{table}\makeatother
            \setlength{\tabcolsep}{1mm}{
                \begin{tabular}{l|c|c|c}
                    \toprule
                    \multirow{1}{*}{\textbf{Metrics} }             & \multirow{1}{*}{\textbf{\oakinkii{}-G.}} & \multirow{1}{*}{\textbf{\oakinki{}-Core}} & \multirow{1}{*}{\textbf{\oakinki{}-Shape}} \\
                    \midrule
                    \textit{Penet. Depth. cm}$\downarrow$          & 0.25                                              & 0.18                                      & 0.11                                       \\
                    \textit{Solid Intsec. Vol. cm$^3$}$\downarrow$ & 0.61                                              & 1.03                                      & 0.62                                       \\
                    \textit{Sim. Disp. Mean cm}$\downarrow$        & 1.83                                              & 0.98                                      & 0.94                                       \\
                    \textit{Sim. Disp. Std cm}$\downarrow$         & 1.16                                              & 1.74                                      & 1.62                                       \\
                    \bottomrule
                \end{tabular}
            }
        }}
    \caption{\textbf{Quality assessment} of \oakinkii{}.}
    \label{tab:dataset_quality}
\end{table}

\section{Tasks and Benchmarks}
\label{sec:app_benchmark}

\subsection{Task-aware Motion Fulfillment}

\label{sec:app_tamf}

\qheading{Train-Val-Test Split} Following the same practice as HMR, we partition the subsets at the sequence level, maintaining the proportion of samples in train/val/test sets at approximately 70\%, 5\%, and 25\%, in alignment with \oakinki{}.

\qheading{Evaluation Metric Details}

\begin{description}[noitemsep,leftmargin=0em,topsep=0em]
    \item[CR, Contact Ratio.]  This metric measures the ratio of the frames within the motion trajectories where the hand-object contact (minimum distance) is within a $\SI{5}{mm}$ threshold.
    \item[SIV, Solid Intersection Volume.]
        This metric measures how much space intersection occurs during estimation.
        We voxelize the object mesh into $100^3$ voxels, and calculate the sum of the voxel volume inside the hand surface.
    \item[PSKL-J, Power Spectrum KL divergence of Joints.]
        This metric reflects the smoothness of the generated motion.
        It measures the acceleration distribution variance between \textbf{predicted} and \textbf{g.t.} joint sequences, reporting results in both directions. 
        We reference our implementation on \cite{wu2022saga,li2024favor}.
        The notable difference is that we use hand joints for measurement, resulting in a distinct range of metric values compared to full-body joints.
    \item[FID, Fr\'{e}chet Inception Distance score.]
        This metric evaluates the realism of the generated motion trajectory. 
        We develop a motion feature extractor based on the transformer encoder architecture.
        The embedding is obtained by appending a trailing token to hand motion trajectories.
        The embedding we use for each motion is of $64$ dimensions.
        The encoder is trained by the classifying motion trajectories into their corresponding categories.
        We apply the encoder to both ground-truth trajectories and generated trajectories and compute Fr\'{e}chet Inception Distance between them for motion realism evaluation.
\end{description}

\section{Application: Complex Task Completion}

\label{sec:app_ctc}

\qheading{Test Scene Generation.}
To evaluate the ability of the oracle-facilitated three-stage method described in the main text to accomplish complex tasks, we derive a set of test environments by perturbing the object positions within the complex scenarios contained in \oakinkii{} dataset without altering the task objectives $\mathbf{text}_\mathrm{goal}$ and the descriptions of the objects' states $\{\mathbf{text}_\mathrm{obj}\}$.

We utilize ground truth annotations to generate variations in the test scenes.
We treat specific object sets as clusters and place them in randomized locations.
Objects within the same cluster share a unified offset to ensure collective randomization. 
This is crucial when groups of objects must maintain coherence in their movements.
The process of randomization comprises four distinct \textit{wander} steps.
This helps prevent obstruction caused by other objects when an object ventures in a random direction.
Hence, each object gains an enhanced opportunity to navigate around other structures within its environment.
Each object's final location results from the cumulative effect of these four \textit{wander} steps.
Within each \textit{wander} step, a maximum of eight iterations are employed for collision prevention between objects by reducing the step length to half. 

\qheading{Prompt Generation.}
We implement Primitive Planning by tweaking the Large Language Model -- GPT-4~\cite{openai2023gpt} in this study -- so that it can generate Python code based on narrative prompts describing the current scenario and task objectives.
The language model's role is to interpret this description, identify key objects involved in the task, determine the appropriate object affordance and trajectory, and generate suitable instances of \textit{Primitives} execution organized into a feasible sequence.

Our approach to overcoming these challenges involves the design of a prompt template that incorporates both scene and task descriptions as referenced earlier. 
This template not only explicates the underlying code framework but also provides a sample of scenario-independent code. 
We further prompt the Language Learning Model (LLM) to produce a code implementation as a response, as opposed to providing an explanatory narrative of coding procedures. 

Concerning the underlying code framework, to ensure robust and coherent code generation, we propose an Entity-Component-System (ECS) architecture.
This structure encourages a decoupling of components, here referred to as data or state, from the system, representing the \textit{Primitives} in our context.
This approach endows us with the capability of generating uniformly styled code implementations, where the layout involves instantiating object entities, loading the affordance as a component, and submitting the \textit{Primitives} to the execution system.

\qheading{Evaluations of Primitive Planning.}
We employ a checker based on the \textit{Primitive} Dependency Graph provided along with the \textit{Complex Tasks} to be planned to benchmark the success rate of the program that is supposed to complete the task target.
We analyze the checker results and observe an overall success rate of 36\% in the generation of Planning codes. 
Concerning the number of \textit{Primitives} incorporated within the \textit{Complex Tasks}, we observed differing success rates. 
Specifically, in those \textit{Complex Tasks} incorporating equal to or less than three \textit{Primitives}, a success rate of 44\% was obtained
Conversely, in the \textit{Complex Tasks} category incorporating between three and five \textit{Primitives}, the success rate dropped to 20\%. 
Notably, no success was recorded in \textit{Complex Tasks} incorporating more than five \textit{Primitives}.
The results demonstrate that in the current setting, the Large Language Model (LLM) is adequate to handle relatively simpler \textit{Complex Tasks}.
However, in contexts of highly complex \textit{Complex Tasks}, the LLM struggles to accurately comprehend the relationships and dependencies between objects' affordances and the corresponding \textit{Primitives}.

\qheading{Demo Planning Result.}
We provide a review of the results of a completed \textit{Complex Task} within one of the constructed test scenes. 
The python programs generated are listed as \cref{listing:ptp_succ}.
This code joins all the relevant objects and their associated affordances, proceeding to execute the \textit{Primitives} in the precise required order. 
We have also included an example of a failed case, presented as \cref{listing:ptp_fail}, which highlights a failure in the execution of \textit{Primitive} Planning.
This failure is characterized by a superfluous \textit{Primitive} that fulfills an unnecessary object affordance that blocks the execution path.

\begin{figure}[t]
    \centering
    \vspace{-0.0em}
    \begin{minipage}{1.00\linewidth}
        \centering
        \includegraphics[trim=000mm 000mm 000mm 000mm, clip=False, width=\linewidth]{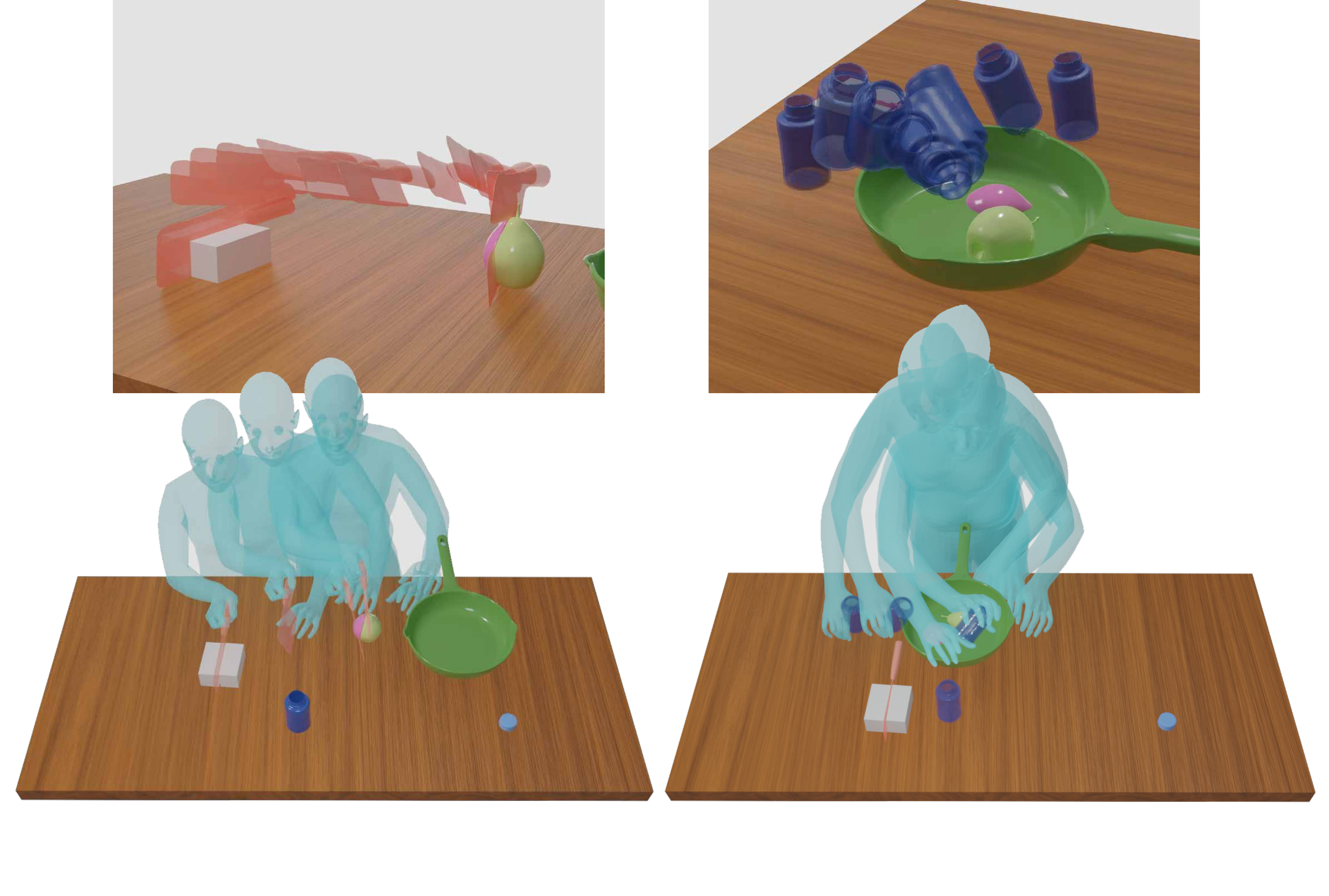}
    \end{minipage}
    \captionsetup{type=figure}
    \captionof{figure}{\cheading{Oracle Trajectories and Motion Generation}
    This figure illustrates the successful \textit{Complex Task} completion of two Primitive Tasks. The top pair of images depict the oracle trajectories, while the bottom pair represents the sequential motion generated.}
    \vspace{-0.0em}
    \label{fig:ctc_infer}
\end{figure}

\vspace{5pt}\noindent\textbf{Alternative Motion Generation for Complex Task Completion.}
In addition to the TaMF-based motion generation approach presented in the main text, we explore an alternative strategy that leverages keyframe generation and motion in-betweening as motion generator for Complex Task Completion.
We adopt GNet and MNet in GOAL \cite{taheri2022goal} and INet in FAVOR \cite{li2024favor}.
These models follow the pattern of first generating hand-object interactions in key frames, and then generating intermediary interaction trajectories within these frames.
The generation contains three stages.
In the first stage, GNet generates static grasps based on the object's initial and terminal poses.
Subsequently, MNet generates motion trajectories to reach the object and retreat from the object.
In the final stage, INet is fed with alternating object poses from the object motion trajectory to generate the in-between motion during the interaction process.
The object oracle trajectories result in a sequence of human body movements depicted in \cref{fig:ctc_infer}. 
The left-hand images of \cref{fig:ctc_infer} illustrate the sequential actions of approaching and utilizing a knife to cut a pear, representing the affordance \textit{cut} associated with the knife under the \textit{Primitive Task} category.
The right-hand images illustrate the sequence of lifting a bottle and pouring its contents into a pan, indicative of the \textit{Primitive Task} affordance, \textit{pour}, as related to the bottle.

\clearpage

\onecolumn
\section{Dataset Inspection}

\label{sec:data_inspect}

\subsection{Task List}

\label{sec:di_tasks}

{
    \scriptsize
    \begin{longtable}[c]{|p{.10\textwidth}p{.30\textwidth}p{.30\textwidth}p{.20\textwidth}|}
        \caption{Collected \textit{Affordances} and Designed \textit{Primitive Tasks}.}
        \label{tab:r_pt_ct}                                                                                                                                                              \\
        \hline
        {\bfseries Scenario} & {\bfseries Affordance}                           & {\bfseries Affordance Instantiation}                   & {\bfseries \textit{Primitive Task}}                      \\
        \hline
        \endhead

        kitchen table        & \texttt{<be rearranged, \_>}                     &                                             & \primitiveid{rearrange}                                  \\
                             & \texttt{<store securely, sth>}                   &                                             &                                                          \\
                             & \qquad \texttt{<contain, sth>}                   &                                             &                                                          \\
                             &                                                  & \texttt{<flow in, sth>}                     & \primitiveid{pour}                                       \\
                             &                                                  & \texttt{<pour, sth>}                        & \primitiveid{pour}                                       \\
                             &                                                  & \texttt{<shake, sth>}                       & \primitiveid{shake}                                      \\
                             & \qquad \texttt{<secure, sth>}                    &                                             &                                                          \\
                             &                                                  & \texttt{<screw into, sth>}                  & \primitiveid{screw}                                      \\
                             &                                                  & \texttt{<unscrew from, sth>}                & \primitiveid{unscrew}                                    \\
                             &                                                  & \texttt{<cap onto, sth>}                    & \primitiveid{cap}                                        \\
                             &                                                  & \texttt{<uncap from, sth>}                  & \primitiveid{uncap}                                      \\
                             & \texttt{<grip, sth>}                             &                                             & \primitiveid{grip}                                       \\
                             & \texttt{<scoop, sth>}                            &                                             & \primitiveid{scoop}                                      \\
                             & \texttt{<scrape, sth>}                           &                                             & \primitiveid{scrape}                                     \\
                             & \texttt{<cut, sth>}                              &                                             & \primitiveid{cut}                                        \\
                             & \texttt{<stir, sth>}                             &                                             & \primitiveid{stir}                                       \\
                             & \texttt{<spread, sth>}                           &                                             & \primitiveid{spread}                                     \\
                             & \texttt{<assemble into, sth>}                    &                                             & \primitiveid{assemble}                                   \\
                             & \texttt{<wipe, sth>}                             &                                             & \primitiveid{wipe}                                       \\
                             & \texttt{<heat with microwave, sth>}              &                                             &                                                          \\
                             & \qquad \texttt{<contain, sth>}                   &                                             &                                                          \\
                             &                                                  & \texttt{<place inside, sth>}                & {\color{gray} \primitiveid{place inside}}                \\
                             &                                                  & \texttt{<take outside, sth>}                & {\color{gray} \primitiveid{take outside}}                \\
                             & \qquad \texttt{<secure, sth>}                    &                                             &                                                          \\
                             &                                                  & \texttt{<shut, sth>}                        & \primitiveid{close gate}                                 \\
                             &                                                  & \texttt{<open, sth>}                        & \primitiveid{open gate}                                  \\
                             & \qquad \texttt{<control, sth>}                   &                                             &                                                          \\
                             &                                                  & \texttt{<be pressed, \_>}                   & {\color{gray}\primitiveid{press button}}                 \\
                             &                                                  & \texttt{<trigger, sth>}                     & {\color{gray}\primitiveid{trigger lever}}                \\
                             & \texttt{<weigh, sth>}                            &                                             &                                                          \\
                             & \qquad \texttt{<support, sth>}                   &                                             & {\color{gray}\primitiveid{place onto}}                   \\
        \hline
        study room table     & \texttt{<be rearranged, \_>}                     &                                             & \primitiveid{rearrange}                                  \\
                             & \texttt{<store securely, sth>}                   &                                             &                                                          \\
                             & \qquad \texttt{<contain, sth>}                   &                                             &                                                          \\
                             &                                                  & \texttt{<place inside, sth>}                &                                                          \\
                             &                                                  & \texttt{<take outside, sth>}                &                                                          \\
                             & \qquad \texttt{<secure, sth>}                    &                                             &                                                          \\
                             &                                                  & \texttt{<cover, sth>}                       & {\color{gray} \primitiveid{put on lid}}                  \\
                             &                                                  & \texttt{<uncover, sth>}                     & {\color{gray} \primitiveid{remove lid}}                  \\
                             &                                                  & \texttt{<shut, sth>}                        & {\color{gray} \primitiveid{pull out drawer}}             \\
                             &                                                  & \texttt{<open, sth>}                        & {\color{gray} \primitiveid{push in drawer}}              \\
                             & \texttt{<illuminate, sth>}                       &                                             &                                                          \\
                             & \texttt{<connect to, sth>}                       &                                             &                                                          \\
                             &                                                  & \texttt{<connect to, power socket>}         & \primitiveid{plug in power plug}                         \\
                             &                                                  & \texttt{<deconnect from, power socket>}     & \primitiveid{remove power plug}                          \\
                             &                                                  & \texttt{<connect to, usb>}                  & \primitiveid{insert usb}                                 \\
                             &                                                  & \texttt{<deconnect from, usb>}              & \primitiveid{remove usb}                                 \\
                             &                                                  & \texttt{<connect to, lightbulb socket>}     & \primitiveid{insert lightbulb}                           \\
                             &                                                  & \texttt{<deconnect from, lightbulb socket>} & \primitiveid{remove lightbulb}                           \\
                             & \texttt{<shear, paper>}                          &                                             &                                                          \\
                             & \texttt{<secure, sth>}                           &                                             &                                                          \\
                             &                                                  & \texttt{<cap, pen tip>}                     & {\color{gray} \primitiveid{cap the pen}}                 \\
                             &                                                  & \texttt{<uncap, pen tip>}                   & {\color{gray} \primitiveid{remove the pen cap}}          \\
                             & \texttt{<write/draw, sth>}                       &                                             & \primitiveid{write on paper}                             \\
                             &                                                  &                                             & \primitiveid{write on whiteboard}                        \\
                             & \texttt{<brush, whiteboard>}                     &                                             & \primitiveid{brush whiteboard}                           \\
                             & \texttt{<be sharpen by, sth>}                    &                                             & \primitiveid{sharpen pencil}                             \\
                             & \texttt{<sharpen, pencil>}                       &                                             & \primitiveid{sharpen pencil}                             \\
                             & \texttt{<staple together, paper>}                &                                             & \primitiveid{staple paper together}                      \\
                             & \texttt{<be written/drawn by, pen/pencil>}       &                                             & \primitiveid{write on paper}                             \\
                             &                                                  &                                             & \primitiveid{write on whiteboard}                        \\
                             & \texttt{<be sheared by, scissors>}               &                                             & \primitiveid{shear paper}                                \\
                             & \texttt{<be stapled together by, stapler>}       &                                             & \primitiveid{staple paper together}                      \\
                             & \texttt{<be turn, \_>}                           &                                             & \primitiveid{close book}                                 \\
                             &                                                  &                                             & \primitiveid{open book}                                  \\
                             & \texttt{<display, sth>}                          &                                             &                                                          \\
                             & \qquad \texttt{<protect, sth>}                   &                                             &                                                          \\
                             &                                                  & \texttt{<open, laptop lid>}                 & \primitiveid{open laptop lid}                            \\
                             &                                                  & \texttt{<close, laptop lid>}                & \primitiveid{close laptop lid}                           \\
                             & \texttt{<control, sth>}                          &                                             & {\color{gray} \primitiveid{use keyboard}}                \\
                             &                                                  &                                             & {\color{gray} \primitiveid{use mouse}}                   \\
                             &                                                  &                                             & {\color{gray} \primitiveid{use gamecontroller}}          \\
                             & \texttt{<cultivate, flowers>}                    &                                             & \primitiveid{put flower into vase}                       \\
                             & \texttt{<be cultivated in, sth>}                 &                                             & \primitiveid{put flower into vase}                       \\
        \hline
        demo chem lab        & \texttt{<be rearranged, \_>}                     &                                             & \primitiveid{rearrange}                                  \\
                             & \texttt{<store securely, experiment substances>} &                                             &                                                          \\
                             & \qquad \texttt{<contain, experiment substances>} &                                             &                                                          \\
                             &                                                  & \texttt{<flow in, experiment substances>}   & \primitiveid{pour in lab}                                \\
                             &                                                  & \texttt{<pour, experiment substances>}      & \primitiveid{pour in lab}                                \\
                             &                                                  & \texttt{<shake, experiment substances>}     & \primitiveid{shake lab container}                        \\
                             & \qquad \texttt{<secure, experiment substances>}  &                                             &                                                          \\
                             &                                                  & \texttt{<screw into, lab container>}        & {\color{gray}\primitiveid{screw}}                        \\
                             &                                                  & \texttt{<unscrew from, lab container>}      & {\color{gray}\primitiveid{unscrew}}                      \\
                             &                                                  & \texttt{<cap onto, lab container>}          & {\color{gray}\primitiveid{cap}}                          \\
                             &                                                  & \texttt{<uncap from, lab container>}        & {\color{gray}\primitiveid{uncap}}                        \\
                             & \texttt{<contain, experiment substances>}        &                                             &                                                          \\
                             &                                                  & \texttt{<flow in, experiment substances>}   & \primitiveid{pour in lab}                                \\
                             &                                                  & \texttt{<pour, experiment substances>}      & \primitiveid{pour in lab}                                \\
                             &                                                  & \texttt{<shake, experiment substances>}     & \primitiveid{shake lab container}                        \\
                             & \texttt{<be heated by, alcohol lamp>}            &                                             & \primitiveid{heat beaker/flask}                        \\
                             &                                                  &                                             & \primitiveid{heat test tube}                             \\
                             & \texttt{<stir, experiment substances>}           &                                             & \primitiveid{stir experiment substances}                 \\
                             & \texttt{<be ignited, \_>}                        &                                             & \primitiveid{ignite alcohol lamp}                        \\
                             & \texttt{<heat, lab container>}                   &                                             & \primitiveid{heat beaker/flask}                        \\
                             &                                                  &                                             & \primitiveid{heat test tube}                             \\
                             & \texttt{<put off, alcohol lamp>}                 &                                             & \primitiveid{put off alcohol lamp}                       \\
                             & \texttt{<ignite, alcohol lamp>}                  &                                             & \primitiveid{ignite alcohol lamp}                        \\
                             & \texttt{<clamp, test tube>}                      &                                             & {\color{gray} \primitiveid{hold test tube} }             \\
                             & \texttt{<conduct heat to, lab container>}        &                                             & \primitiveid{place asbestos mesh}                        \\
                             & \texttt{<support, lab container>}                &                                             & \primitiveid{place asbestos mesh}                        \\
        \hline
        bathroom table       & \texttt{<be rearranged, \_>}                     &                                             & \primitiveid{rearrange}                                  \\
                             & \texttt{<contain, sth>}                          &                                             &                                                          \\
                             &                                                  & \texttt{<squeeze out, sth>}                 & {\color{gray} \primitiveid{squeeze tooth paste} }        \\
                             & \texttt{<secure, sth>}                           &                                             &                                                          \\
                             &                                                  & \texttt{<shut, sth>}                        & {\color{gray} \primitiveid{flip open tooth paste cap} }  \\
                             &                                                  & \texttt{<open, sth>}                        & {\color{gray} \primitiveid{flip close tooth paste cap} } \\
        \hline
        \caption{Collected \textit{Affordances} and Designed \textit{Primitive Tasks}.
            The first column records the manipulation scenarios.
            The second column lists collected affordances of object instances and parts. The affordances of object parts are indented below their parent instance-level affordance.
            The third column lists the instantiations of object affordances. These instantiations are bound to certain object attributes, \eg \texttt{<screw, sth>} is bound to actual screws on the bottle's opening and cap.
            The fourth column lists the designed \textit{Primitives} corresponding to the affordances.
            Some \textit{Primitives} are set to gray for these \textit{Primitives} are difficult to demonstrate and capture in individual.
            Demonstrations of these tasks are embedded within \textit{Complex Task} demonstrations.
        }
    \end{longtable}
}

{
    \footnotesize
    \begin{longtable}[c]{|p{.15\textwidth}p{.8\textwidth}|}
        \hline
        {\bfseries Scenario} & {\bfseries \textit{Complex Task}} \\
        \hline
        \endhead
        kitchen table        &
        heat with microwave oven;
        weigh with scale;
        scoop and pour;
        scoop and grip;
        scoop and wipe;
        scoop and scrape;
        pour and stir;
        grip and pour;
        pour and arrange;
        weigh with scale and pour;
        pour and scrape;
        grip and arrange;
        weigh with scale and grip;
        grip and wipe;
        pour and grip;
        clean the kitchen table;
        prepare a cup of hot sweet drink;
        prepare a bowl of hot soup with salt;
        prepare a cup of hot sweet fruit tea;
        prepare a chilled apple platter;
        prepare a savory fruit salad;
        prepare a baked sweet donut with sauce;
        prepare a baked sweet donut with apple slices and jam;
        prepare a savory baked sweet donut;
        prepare a cheese-baked sweet donut with tomato sauce;
        prepare savory baked apple slices with cheese;
        prepare a chilled fruit platter;
        make a baked sandwich with a filling of donut and salt;
        make a sandwich with a filling of tomato sauce and sugar;
        make a sandwich with a filling of apple slices and donut, adding tomato sauce, mustard sauce, salt, and sugar;
        make a baked sandwich with a filling of cheese and donut, adding tomato sauce;
        scoop, unscrew, pour, and screw;
        grip and scoop;
        scoop and scoop;
        scoop and arrange;
        weigh with scale and scoop;
        cut and scoop;
        cut and pour;
        grip and stir;
        grip and scrape;
        cut and grip;
        grip and assemble;
        stir and arrange;
        stir and scrape;
        scrape and arrange;
        weigh with scale and assemble;
        unscrew and pour;
        pour and screw;
        uncap and scrape;
        scrape and cap;
        uncap, scrape, and cap;
        scrape and assemble;
        assemble and arrange;
        unscrew and heat with microwave;
        pour and heat with microwave;
        heat with microwave and pour;
        heat with microwave and stir;
        heat with microwave and assemble;
        cut and heat with microwave;
        uncap, pour, and cap;
        prepare a cup of chilled green tea;
        prepare a cup of apple green tea;
        prepare a cup of mixed flavor fruit juice;
        prepare a cup of savory fruit juice milk tea;
        prepare a cup of pear milk tea with fruit jam;
        prepare a cup of savory honey fruit juice milk tea;
        prepare a cup of savory strawberry orange juice mixed with milk;
        uncap and scoop;
        prepare a cup of wine;
        prepare a cup of milk tea;
        prepare a cup of chilled fruit tea;
        prepare a cup of honey coffee;
        prepare a cup of chilled juice milk with jam;
        prepare a cup of chilled sweet milk tea;                 \\
        \hline
        study room table     &
        put into box;
        take out of box;
        put into drawer;
        take out of drawer;
        ready the laptop on the desktop for work;
        tidy up the desktop with the laptop after work;
        ready the laptop on the desktop for entertainment;
        illuminate the desktop;
        sharpen the pencil and write;
        tidy up the desktop with the laptop after entertainment;
        tidy up the desktop after paper-cutting;
        write and bind the paper;
        design and cut out rectangle shape on the paper;
        press button and open laptop;
        press button and close laptop;
        press button and put into box;
        press button and take out of box;
        press button and remove power plug;
        insert usb and plug in power plug;
        tidy up the desktop after writing;
        plug in power plug and press button;
        design and cut out flower shape on the paper;
        design and draw on the paper;
        ready the laptop and the lamp on the desktop for work;
        design, write and bind the paper;
        ready the desktop for drawing;
        take out of drawer, insert usb, and open laptop;
        put into drawer and put into box;
        put into drawer, put into box, and close laptop;
        remove usb, close laptop, and put into drawer;
        tidy up the desktop after drawing;
        cut and bind paper;
        ready the laptop and the lamp on the desktop for entertainment;
        design, draw and cut out flower shape on the paper;
        design, draw and bind the paper;
        tidy up the desktop after binding paper;                 \\
        \hline
        demo chem lab        &
        transfer and heat liquid in beaker;
        transfer and heat liquid in conical flask;
        heat liquid in beaker and transfer liquid;
        heat liquid in conical flask and transfer liquid;
        transfer and heat liquid in beaker and transfer liquid out;
        heat liquid in test tube and transfer liquid;
        prepare solution through heating;
        mix liquid;
        pour in lab and shake in lab;
        pour in lab and pour in lab;
        pour in lab and heat test tube;
        pour in lab and light lamp;
        stir in lab and pour in lab;
        stir in lab and heat beaker;
        shake in lab and pour in lab;
        shake in lab and heat test tube;
        shake in lab and heat beaker;
        heat beaker and put off lamp;
        heat test tube and put off lamp;
        light lamp and put off lamp;
        put off lamp and pour in lab;
        put off lamp and stir in lab;
        put off lamp and shake in lab;
        heat beaker and stir in lab;
        stack mesh and heat beaker;
        light lamp and heat beaker;
        light lamp and heat test tube;
        pour in lab, shake in lab, and heat test tube;
        stir in lab, pour in lab, and shake in lab;
        light lamp, heat beaker, and put off lamp;
        light lamp, heat test tube, and put off lamp;
        stack mesh, light lamp, and heat beaker;
        light lamp, heat beaker, and stir in lab;
        light lamp, heat test tube, and shake in lab;
        pour in lab, pour in lab, and pour in lab;
        pour in lab, shake in lab, and pour in lab;
        stir in lab, stack mesh, and heat beaker;
        shake in lab, pour in lab, and heat test tube;
        shake in lab, heat test tube, and pour in lab;
        heat beaker, stir in lab, and pour in lab;
        heat beaker, put off lamp, and pour in lab;
        heat beaker, put off lamp, and stir in lab;
        heat test tube, put off lamp, and shake in lab;
        pour in lab, stir in lab, and pour in lab;
        pour in lab, pour in lab, pour in lab, pour in lab, and pour in lab;
        stir and transfer liquid;
        heat liquid in beaker;
        heat liquid in test tube;
        put off lamp, pour in lab, and shake in lab;
        put off lamp, stir in lab, and pour in lab;
        prepare solution in beaker;                              \\
        \hline
        bathroom table       &
        squeeze tooth paste tube to tooth brush;
        squeeze tooth paste and stack tooth brush;
        prepare for teeth brushing.                              \\
        \hline
        \caption{Recorded \textit{Complex Tasks}.
            We list the names of the recorded \textit{Complex Tasks} here.
        }
        \label{tab:r_ct}
    \end{longtable}
}

\subsection{Visualization}
\label{sec:di_viz}

\begin{figure*}[t]
    \centering
    \begin{minipage}{1.00\linewidth}
        \centering
        \includegraphics[trim=000mm 000mm 000mm 000mm, clip=False, width=\linewidth]{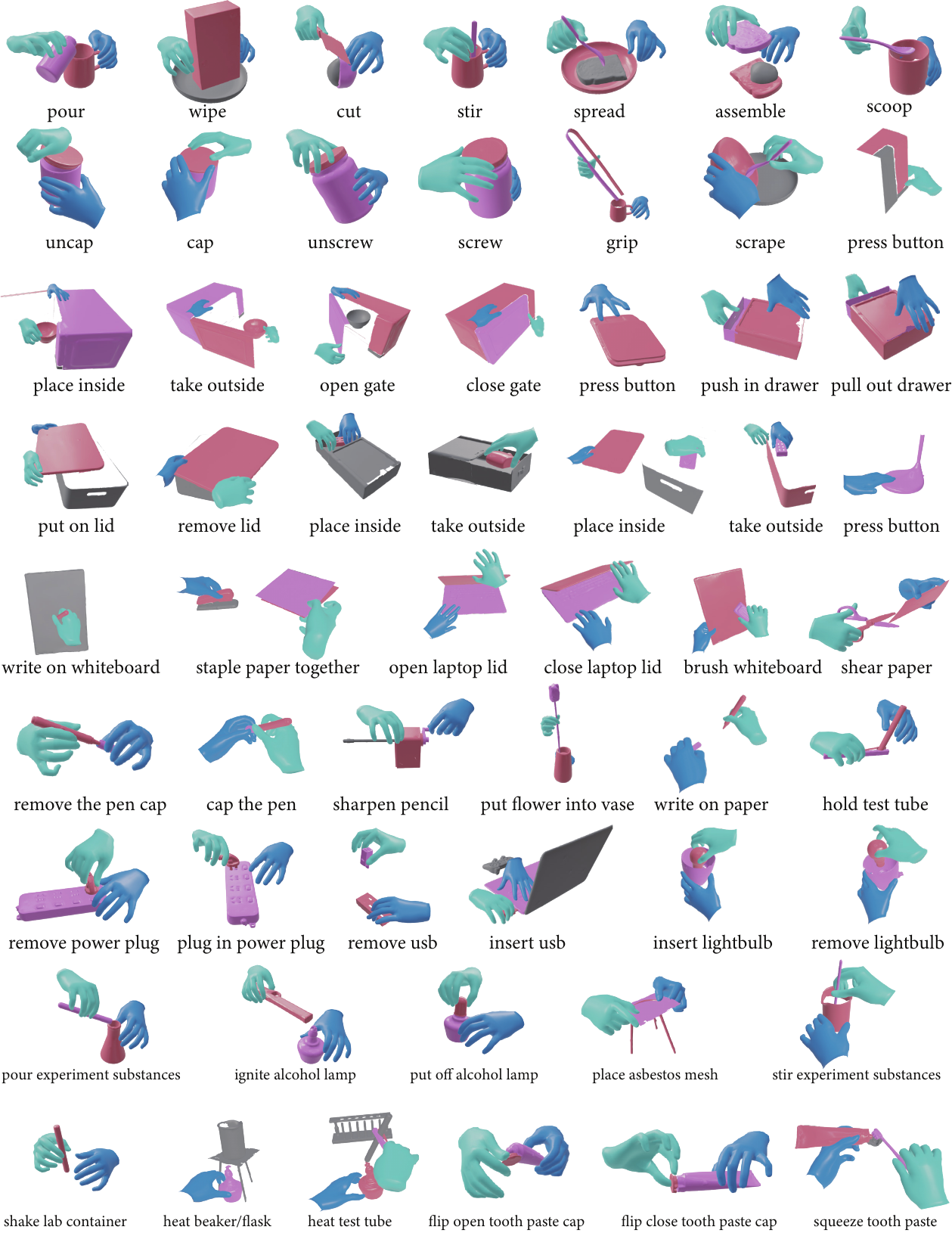}
    \end{minipage}
    \captionsetup{type=figure}
    \captionof{figure}{\cheading{\textit{Primitives} visualization.}}
    \vspace{0.0em}
    \label{fig:primitive_combo_viz}
\end{figure*}

\begin{figure*}[t]
    \centering
    \begin{minipage}{1.00\linewidth}
        \centering
        \includegraphics[trim=000mm 000mm 000mm 000mm, clip=False, width=\linewidth]{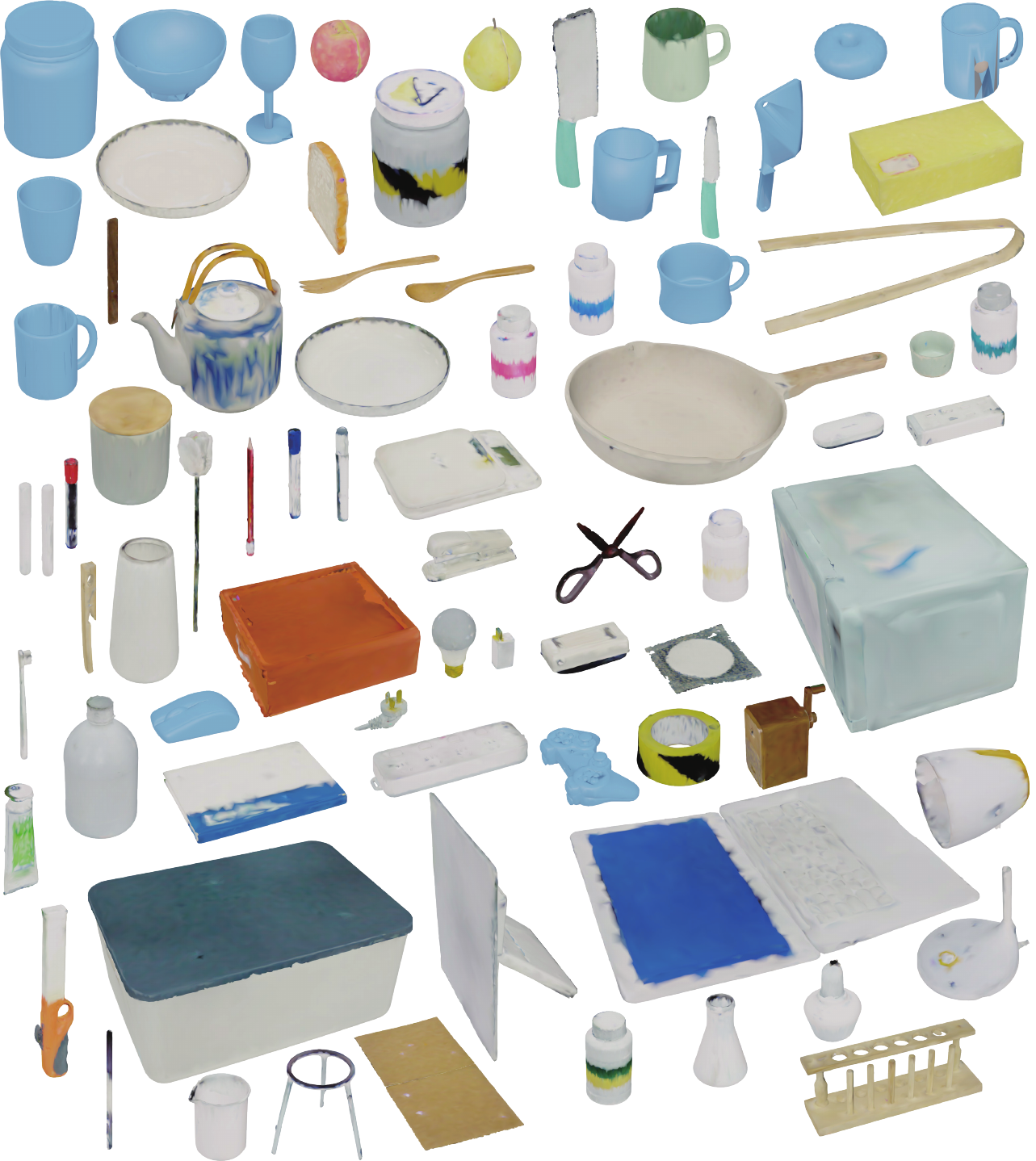}
    \end{minipage}
    \captionsetup{type=figure}
    \captionof{figure}{\cheading{Object visualization.}}
    \vspace{0.0em}
    \label{fig:obj_combo_viz}
\end{figure*}

\begin{figure*}[!htb]
    \lstset{
        numbers=left, numberstyle=\tiny, stepnumber=1, numbersep=5pt,
    }
    \begin{lstlisting}[
    morekeywords={entities,Entity,components,Component,systems,System,oracle,generate},
    keywordstyle=\bfseries,
]
You are a python programming expert and you are asked to finish a certain bimanual robotics task.

Scene Description:
{scene_desc}

Task Description: 
{task_desc}

The Structure of the code is a ECS architecture defined as 
ECS File

The entities are defined as
Entity File

The components are defined as
Component File

The systems are defined as
System File

The task is to finish the methods called "exec_task" in this class {read_from_file(scene_path)}

You need to query the raw 3d object from the scene which contains object name as keys in scene description and you can use them to query different type of information from the scene. Build them into Objects and instantiate PrimitiveTasks to finish the job.
Leave the objects not mentioned in the task description as they are.
Avoid using any methods with underscore prefix. Explicitly specify the keyword arguments instead of using **kwargs. 
For example:

an_object = Object('object_name', 
    geometry=self.query_geometry_info('object_name'), 
    ...
    )

a_primitive_task = PrimitiveTask(an_object.affordance.get_primitive_task_info('primitive_task_name'))

a_primitive_task.execute(src_object=an_object, tgt_object=another_object, trajectory=oracle.generate(a_primitive_task))

Explanation of the code is unnecessary. Putting everything in method implementation would be admired. Respond with exec_task(self) itself.
\end{lstlisting}
    \captionof{lstlisting}{Prompt Template.
        \textbf{Entity} marks object instances present in the scene.
        \textbf{Component} accommodates information of objects' initial status and affordances.
        \textbf{System} accommodates the interface of motion generators for \textit{Primitives}.
        The example code incorporated in the prompt demonstrates the interface of \textbf{oracle queries} for object motion trajectories.
    }
\end{figure*}

\begin{figure*}[!htb]
    \lstset{
        numbers=left, numberstyle=\tiny, stepnumber=1, numbersep=5pt,
    }
    \begin{lstlisting}[language=Python]
# Task
# The task is to cut the pear into pieces which to be put in the pan and add some season.

# Scene
# In a kitchen, on a table, these objects are placed: [knife, pear, bottle, pan]
def exec_task(self):
    # making an object for knife
    knife = Object('knife', 
        geometry=self.query_geometry_info('knife'), 
        state=self.query_state_info('knife'),
        affordance=self.query_affordance_info('knife')
    )

    # making an object for pear
    pear = Object('pear', 
        geometry=self.query_geometry_info('pear'), 
        state=self.query_state_info('pear'),
        affordance=self.query_affordance_info('pear')
    )

    # making an object for pan
    pan = Object('pan', 
        geometry=self.query_geometry_info('pan'), 
        state=self.query_state_info('pan'),
        affordance=self.query_affordance_info('pan')
    )

    # making an object for bottle
    bottle = Object('bottle', 
        geometry=self.query_geometry_info('bottle'), 
        state=self.query_state_info('bottle'),
        affordance=self.query_affordance_info('bottle')
    )

    # creating primitive task for cutting
    cut_pear = PrimitiveTask(knife.affordance.get_primitive_task_info('cut'))
    cut_pear.execute(src_object=knife, tgt_object=pear, trajectory=oracle.generate(cut_pear))

    # creating primitive task for placing pear inside the pan
    put_task = PrimitiveTask(pear.affordance.get_primitive_task_info('put_inside'))
    put_task.execute(src_object=pear, tgt_object=pan)

    # create primitive task for adding season
    add_season = PrimitiveTask(bottle.affordance.get_primitive_task_info('pour'))
    add_season.execute(src_object=bottle, tgt_object=pan, trajectory=oracle.generate(add_season))
\end{lstlisting}
    \captionof{lstlisting}{Primitive Task Planning Success Example.}
    \label{listing:ptp_succ}
\end{figure*}

\begin{figure*}[!htb]
    \lstset{
        numbers=left, numberstyle=\tiny, stepnumber=1, numbersep=5pt,
    }
    \begin{lstlisting}[language=Python]
# Task
# The task is to light the alcohol lamp to heat the compound in the beaker. The lamp is capped at the beginning. Put the beaker on the support with mesh when heating

# Scene
# In a chemistry laboratory, on a table, these objects are placed: [lighter, alcohol_lamp, beaker, heat_support, mesh]
def exec_task(self):
    # Uncap
    alcohol_lamp = Object('alcohol_lamp', 
        geometry=self.query_geometry_info('alcohol_lamp'), 
        affordance=self.query_affordance_info('alcohol_lamp'), 
        state=self.query_state_info('alcohol_lamp')
    )
    uncap_alcohol_lamp_task = PrimitiveTask(alcohol_lamp.affordance.get_primitive_task_info('uncap'))
    uncap_alcohol_lamp_task.execute(src_object=alcohol_lamp, trajectory=oracle.generate(uncap_alcohol_lamp_task))

    # Light
    lighter = Object('lighter', 
        geometry=self.query_geometry_info('lighter'), 
        affordance=self.query_affordance_info('lighter'), 
        state=self.query_state_info('lighter')
    )
    light_alcohol_lamp_task = PrimitiveTask(lighter.affordance.get_primitive_task_info('light_lamp'))
    light_alcohol_lamp_task.execute(src_object=lighter, tgt_object=alcohol_lamp, trajectory=oracle.generate(light_alcohol_lamp_task))

    # Arrange (put on the mesh support)
    beaker = Object('beaker', 
         geometry=self.query_geometry_info('beaker'), 
         affordance=self.query_affordance_info('beaker'), 
         state=self.query_state_info('beaker')
    )
    heat_support = Object('heat_support', 
        geometry=self.query_geometry_info('heat_support'), 
        affordance=self.query_affordance_info('heat_support'), 
        state=self.query_state_info('heat_support')
    )
    mesh = Object('mesh', 
         geometry=self.query_geometry_info('mesh'), 
         affordance=self.query_affordance_info('mesh'), 
         state=self.query_state_info('mesh')
    )
    put_beaker_on_mesh_task = PrimitiveTask(beaker.affordance.get_primitive_task_info('arrange'))
    put_beaker_on_mesh_task.execute(src_object=beaker, tgt_object=heat_support, trajectory=oracle.generate(put_beaker_on_mesh_task))

    # Stack mesh to heat
    stack_mesh_task = PrimitiveTask(mesh.affordance.get_primitive_task_info('stack_mesh'))
    stack_mesh_task.execute(src_object=mesh, tgt_object=beaker, trajectory=oracle.generate(stack_mesh_task))

    # Heat
    heat_beaker = PrimitiveTask(alcohol_lamp.affordance.get_primitive_task_info('heat_beaker'))
    heat_beaker.execute(src_object=alcohol_lamp, tgt_object=beaker, trajectory=oracle.generate(heat_beaker))

\end{lstlisting}
    \captionof{lstlisting}{Primitive Task Planning Fail Example. 
    One violation of dependency occurs (line 41): the extra \primitiveid{place onto} primitive erroneously positions the beaker upon the support prior to the asbestos mesh's placement on the support. 
    This blocks the correct execution path that requires placing the asbestos mesh before heating the beaker.
}
    \label{listing:ptp_fail}
\end{figure*}

\twocolumn

\end{appendices}

\end{document}